\documentclass[10pt,journal,compsoc]{IEEEtran}
\ifCLASSOPTIONcompsoc
  \usepackage[nocompress]{cite}
\else
  \usepackage{cite}
\fi
  \usepackage{graphicx}
\usepackage{amsmath}
\usepackage{amssymb}
\usepackage{algorithmic}

\usepackage{array}

\ifCLASSOPTIONcompsoc
  \usepackage[caption=false,font=footnotesize,labelfont=sf,textfont=sf]{subfig}
\else
  \usepackage[caption=false,font=footnotesize]{subfig}
\fi

\usepackage{stfloats}
\usepackage{url}

\newenvironment{packed_itemize}{
\vspace{-0.15cm}\begin{itemize}
  \setlength{\itemsep}{1pt}
  \setlength{\parskip}{0pt}
  \setlength{\parsep}{0pt}
}{\end{itemize}}

\newenvironment{packed_enumerate}{
\begin{enumerate}
  \setlength{\itemsep}{1pt}
  \setlength{\parskip}{0pt}
  \setlength{\parsep}{0pt}
}{\end{enumerate}}

\newcommand{\etal}{\emph{et al.}}
\newcommand{\eg}{\emph{e.g.}}
\newcommand{\ie}{\emph{i.e.}}

\DeclareMathOperator*{\argmax}{argmax}

\DeclareMathOperator{\BigO}{O}
\newcommand{\trans}[1]{{#1}^{\ensuremath{\mathsf{T}}}} 

\usepackage[breaklinks=true,colorlinks,bookmarks=false]{hyperref}

\begin{document}

\title{Deep Imbalanced Learning for Face Recognition and Attribute Prediction}

\author{Chen~Huang,
        Yining~Li,
        Chen~Change~Loy,~\IEEEmembership{Senior Member,~IEEE}
        and~Xiaoou~Tang,~\IEEEmembership{Fellow,~IEEE}
\IEEEcompsocitemizethanks{
\IEEEcompsocthanksitem C. Huang was with the Robotics Institute, Carnegie Mellon University, Pittsburgh, PA, 15213.
\protect\\
E-mail: chenh2@andrew.cmu.edu
\IEEEcompsocthanksitem C. C. Loy is with the School of Computer Science and Engineering, National Technological University, Singapore.
\protect\\
E-mail: ccloy@ntu.edu.sg
\IEEEcompsocthanksitem Y. Li and X. Tang are with the Department of Information Engineering, The Chinese University of Hong Kong.
\protect\\
E-mail: \{ly015,xtang\}@ie.cuhk.edu.hk}

}

%



\IEEEtitleabstractindextext{%
\begin{abstract}
Data for face analysis often exhibit highly-skewed class distribution, \ie,~most data belong to a few majority classes, while the minority classes only contain a scarce amount of instances. To mitigate this issue, contemporary deep learning methods typically follow classic strategies such as class re-sampling or cost-sensitive training. In this paper, we conduct extensive and systematic experiments to validate the effectiveness of these classic schemes for representation learning on class-imbalanced data. We further demonstrate that more discriminative deep representation can be learned by enforcing a deep network to maintain inter-cluster margins both within and between classes. This tight constraint effectively reduces the class imbalance inherent in the local data neighborhood, thus carving much more balanced class boundaries locally. We show that it is easy to deploy angular margins between the cluster distributions on a hypersphere manifold. Such learned Cluster-based Large Margin Local Embedding (CLMLE), when combined with a simple \textit{k}-nearest cluster algorithm, shows significant improvements in accuracy over existing methods on both face recognition and face attribute prediction tasks that exhibit imbalanced class distribution.
\end{abstract}

\begin{IEEEkeywords}
Imbalanced Learning, Deep Convolutional Neural Networks, Face Recognition, Attribute Prediction
\end{IEEEkeywords}}

\maketitle
\IEEEdisplaynontitleabstractindextext

%

\IEEEraisesectionheading{\section{Introduction}\label{sec:introduction}}

\IEEEPARstart{M}{any} data in face analysis domain naturally exhibit imbalance in their class distribution. For instance, the numbers of positive and negative face pairs in face verification~\cite{LFWTech,kemelmacher2016megaface} are highly skewed since it is easier to obtain face images with different identities (negative) than faces with matched identity (positive) during data collection. For face attribute prediction~\cite{kumar2011describable}, it is comparatively easy to find persons with ``normal-sized nose'' attribute from web images than that of ``big-nose''. Such face recognition and attribute prediction problems provide perfect testbeds for studying generic imbalanced learning algorithms, either under closed- or open-set protocol~\cite{Taigman14}. Indeed, without handling the imbalance issue conventional methods tend to be biased toward the majority class with poor accuracy for the minority class~\cite{He09,He2013}.

Deep representation learning has recently achieved great success due to its high learning capacity, but still cannot escape from the negative impact of imbalanced data. To counter such negative effect, one often chooses from a few available options, which have been extensively studied in the past~\cite{He2013,Chawla02,KRAWCZYK2014554,Drummond03,Han05,He09,Maciejewski11,Tang09,Ting00,Chen16,He08adasyn,Zhou2006AAAI}. The \textit{first option} is re-sampling, which aims to balance the class priors by under-sampling the majority class or over-sampling the minority class (or both). For instance, Oquab~\etal~\cite{oquab2014learning} resampled the number of foreground and background image patches for learning a convolutional neural network (CNN) for object classification. The \textit{second option} is cost-sensitive learning, which assigns higher misclassification costs to the minority class than to the majority. Caesar~\etal~\cite{CaesarUF15} proposed to calibrate an ensemble of SVMs with inverse class frequencies as costs to combat class imbalance in semantic segmentation. Similarly in deep CNNs, the loss function is rescaled with the inverse~\cite{MostajabiYS15}, relative~\cite{Rudd2016} and median~\cite{Eigen2015} class frequencies, respectively for semantic segmentation, face attribute prediction and multi-task scene understanding. For image edge detection~\cite{Shen15}, the softmax loss of CNN is regularized with equal weights for the positive and negative edge classes. An alternative~\cite{BuloNK17} goes beyond conventional cost-sensitive strategies by re-weighting the contributions of spatial image pixels based on their actual observed losses for semantic segmentation.

Can these methods help the deep face recognition and attribute prediction tasks where data imbalance is barely handled? Are these methods the most effective way to deal with data imbalance in the context of deep representation learning? The aforementioned options are well studied for the `shallow' model~\cite{Erhan10} but their implications have not yet been systematically studied for deep representation learning. Importantly, such schemes are well-known for some inherent limitations. For instance, over-sampling can easily introduce undesirable noise with increased computational cost and overfitting risk. Under-sampling is often preferred~\cite{Drummond03} but it may remove valuable information. Cost-sensitive learning approaches often design costs using heuristics or static class label statistics.

Such nuisance factors can be equally applicable to the recent deep imbalanced learning methods based on such common schemes. Methods in this line~\cite{Jeatrakul10,Khan18,Castro13,zhou2006training} all fail to provide noticeable improvements in performance. Two advances~\cite{NIPS2017_7278,Dong_2017_ICCV} excel by providing new insights. Wang~\etal~\cite{NIPS2017_7278} proposed a meta-network to transfer knowledge (model parameters) from the majority class to minority class. Thus the model dynamics between \textit{many-shot} and \textit{few-shot} models is learned, achieving superior classification results on existing imbalanced datasets like ImageNet~\cite{Russakovsky2015}. Dong~\etal~\cite{Dong_2017_ICCV} proposed a Class Rectification Loss (CRL) that can further handle imbalanced multi-label attributes and is more related to our goal. CRL performs hard mining for the minority attribute classes in each batch, and enforces their feature constraints to rectify the learning bias of the conventional cross entropy loss. However, regularization for only minority class cannot guarantee equal learning for all classes, and the learning quality and speed can be hindered by the hard mining online that sees one batch at a time, where a global characterization of feature space is lacking for correct regularization.

In this paper, we wish to investigate a more effective approach for deep imbalanced learning. We show its important applications to face recognition\footnote{Face recognition can be categorized as face identification (\ie,~classify one face to a specific identity) and face verification (\ie,~determine whether a pair of faces belong to the same identity).} and attribute prediction from ubiquitously imbalanced datasets. Note such tasks can be evaluated under either closed- or open-set protocol. The open-set protocol is harder since the testing classes may be unseen from the training classes. It usually requires discriminative feature representations with built-in large margins, which are embodied in our approach.

\begin{figure}[t]
\begin{center}
\includegraphics[width=0.75\linewidth]{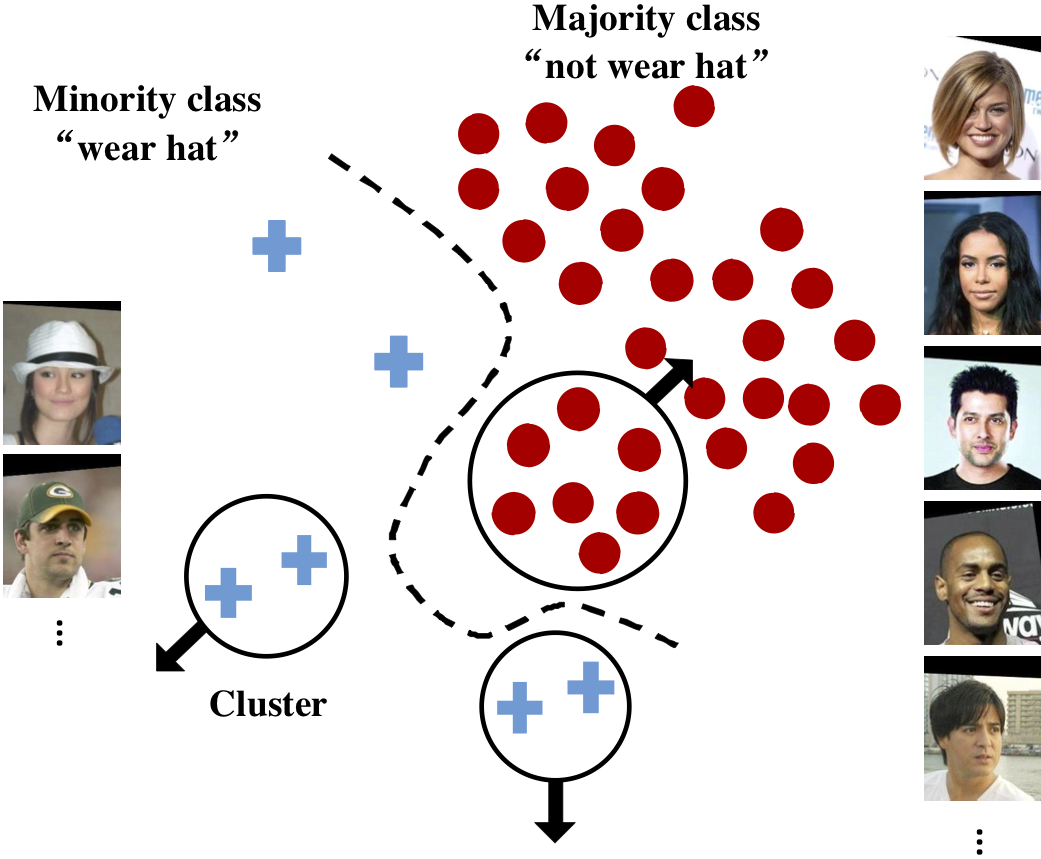}
\end{center}
\vspace{-1.5em}
\caption{Example of class imbalance for the binary face attribute ``wear hat''. Our method aims to separate the cluster distributions both within and between classes. This effectively reduces the class imbalance in local neighborhoods and forms balanced local class boundaries that are insensitive to the imbalanced size of remaining class samples.}
\label{fig1}
\end{figure}

Our method is motivated by the observation that the minority class often contains very few instances with high degree of visual variability. The scarcity and high variability make the genuine neighborhood of these instances easy to be invaded by other imposter nearest neighbors\footnote{An imposter neighbor of a data point $x_i$ is another data point $x_j$ with a different class label, $y_i \neq y_j$.}. Such invasion will confuse the underlying class boundaries formed by either a local or global classifier. To this end, we propose to learn an embedding $f(x) \in \mathbb{R}^d$ with a CNN to ameliorate the invasion. The CNN is trained with a maintained index of clusters for each class, which we will update continuously throughout training. Our objective, then, would enforce margins between hard-mined clusters in the local neighborhood from both the same and different classes. Such margins introduce a tight constraint for reducing local data imbalance, leading to much more balanced class boundaries locally (Fig.~\ref{fig1}). We demonstrate the margins can be deterministically derived on a hypersphere feature space. We also study the effectiveness of classic schemes of re-sampling and cost-sensitive learning in our context.

Using the learned feature representation, we show the evaluation can be simply done by a soft \textit{k}-nearest-cluster metric which is consistent with our learning objective. The proposed approach, called \textit{Cluster-based Large Margin Local Embedding} (CLMLE), achieves the new state-of-the-art performance on several face recognition datasets using only small training data. CLMLE also drastically outperforms the standard softmax and triplet losses and surpasses the recent imbalanced learning methods. For face attribute prediction, CLMLE achieves superior performance measured by the balanced accuracy across multiple attributes.

A preliminary version of this work has been published in~\cite{huang2016lmle}. This work extends the initial method LMLE~\cite{huang2016lmle} in significant ways. (1) Sampling of quintuplets (composed of 5 data points) in LMLE is generalized to the sampling of entire clusters. This alleviates the training inefficiencies of LMLE due to the exponential growth of quintuplet number. Also, penalizing the overlap between cluster distributions is much more coherent than penalizing individual quintuplets or triplets, leading to faster and better convergence than LMLE and triplet loss~\cite{Schroff15}. (2) A new CLMLE loss function is proposed, with customized cluster re-sampling and cost-sensitive learning techniques. (3) We design angular margins to be enforced between the involved cluster distributions. This is more natural than enforcing Euclidean distance on a hypersphere manifold as in LMLE. (4) The effectiveness of CLMLE is validated in the imbalanced tasks of not only face attribute prediction but also face recognition, where the additional open-set scenario demonstrates the superior generalization ability of CLMLE.

\section{Related Work}
Previous efforts to tackle the class imbalance problem can be mainly divided into two groups: data re-sampling~\cite{He2013,Chawla02,Drummond03,Han05,He09,Maciejewski11,He08adasyn} and cost-sensitive learning~\cite{KRAWCZYK2014554,Tang09,Ting00,Zhou2006AAAI,Chen16}. The former group aims to alter the training data distribution to learn equally good classifiers for all classes, usually by random under-sampling and over-sampling techniques. The latter group, instead of manipulating samples at the data level, operates at the algorithmic level by adjusting misclassification costs. A comprehensive literature survey can be found in~\cite{He09,He2013}.

A well-known issue with replication-based random over-sampling is its tendency to overfit. More radically, it does not increase any information and fails in solving the fundamental ``lack of data'' problem for the minority class. To this end, SMOTE~\cite{Chawla02} creates new non-replicated examples by interpolating neighboring minority class instances. Several variants of SMOTE~\cite{Han05,Maciejewski11,He08adasyn} followed for improvements. However, their broadened decision regions are still error-prone by synthesizing borderline examples. Therefore under-sampling is often preferred to over-sampling~\cite{Drummond03}, although potentially valuable information may be removed. Cost-sensitive methods avoid these issues by directly imposing heavier cost on misclassifying the minority class. Currently, how to determine the cost representation is still an open problem. Commonly-agreed practices include using the inverse class frequency or pre-defined misclassification costs, and they are typically applied to SVMs~\cite{Tang09} and decision trees~\cite{Zhou2006AAAI}. Boosting~\cite{Ting00} offers another natural way to embed the costs in example weights. Many other methods follow this philosophy of designing classifier ensemble (\eg,~\cite{Chen16,KRAWCZYK2014554}) to combat imbalance. In~\cite{Chen16,KRAWCZYK2014554}, the authors combined cost sensitivity with bagging which is less vulnerable to noise than boosting, and generated cost-sensitive version of random forests.

\noindent
{\bf Deep imbalanced learning.} To our knowledge, only few works~\cite{Jeatrakul10,Khan18,Castro13,zhou2006training,Wang16IJCNN,NG2016875,NIPS2017_7278,Dong_2017_ICCV} approach imbalanced learning via deep models. Jeatrakul~\etal~\cite{Jeatrakul10} treated the Complementary Neural Network as an under-sampling technique, and combined it with SMOTE-based over-sampling to re-balance data. Zhou and Liu~\cite{zhou2006training} studied data resampling for training cost-sensitive neural networks. In~\cite{Khan18,Castro13}, the cost-sensitive deep features and the cost parameter are jointly optimized. All these works can be seen as direct ``deep'' extensions of traditional imbalanced learning techniques. Alternatives~\cite{Wang16IJCNN,NG2016875} simply tune the networks to maximize a class-balanced accuracy measure. More recently, Ren~\etal~\cite{RenZYU18} proposed to reweight batch samples based on their gradient directions online, which can combat imbalance. However, a clean unbiased validation set is needed to represent the target distribution. Liu~\etal~\cite{LiuNIPS18} proposed to maximize the hyperspherical margin regardless of class imbalance, while Wang~\etal~\cite{NIPS2017_7278} proposed a meta-learning approach that transfers model parameters from the majority to minority class. Both methods achieve good classification results on imbalanced datasets.


Unfortunately, none of the above works takes into account the data structure of imbalanced classes which helps learning. One exception is the Class Rectification Loss (CRL) in~\cite{Dong_2017_ICCV}, where ``hard'' minority classes are searched in each batch and are regularized in feature space to rectify the learning bias of conventional loss,~\eg,~cross entropy. However, feature regularization for only minority class cannot guarantee equal learning for all classes. We propose here a ``structure-aware'' approach by enforcing large margins between intra-class and inter-class clusters. This way, balanced class boundaries can be equally drawn for \textit{every} class from its involved local clusters. We show the clusters provide a global characterization of class distributions, leading to faster and better convergence than purely online methods like CRL~\cite{Dong_2017_ICCV} where the global information is missing. We also show our cluster separation rule can be easily applied to both training and testing, where the clustering process only incurs negligible computational cost during training.

\noindent
{\bf Deep face recognition.} Softmax loss has been pioneering effective CNN models for deep face recognition~\cite{LFWTech,kemelmacher2016megaface}, including recognition under the open-set protocol~\cite{Taigman14}. However, open-set recognition, unlike the closed-set one, cannot be addressed as a classification problem of known face identities as in softmax. Open-set scenario usually requires more discriminatively learned features with built-in margin. Recent L-Softmax loss~\cite{Liu2016}, A-Softmax loss~\cite{liu2017sphereface}, and Large Margin Cosine Loss (LMCL)~\cite{2018Wang} generalize softmax by enforcing large angular margin between classes to enhance feature discrimination. Other works adopt ideas from metric learning, and combine softmax with contrastive~\cite{Yi14,SunWT15,YiLLL14a}, center loss~\cite{wen2016} or marginal loss~\cite{Deng_8014985} for improvements. Another popular choice is the triplet loss~\cite{Schroff15,LiuDBH15}, which leads to state-of-the-art performance. The recent methods that handle class imbalance augment the minority classes in the image space~\cite{2016_dowe} and feature space~\cite{2018arXiv180309014Y}, respectively. Others align the feature centers~\cite{Wu_2017} or weight norms~\cite{Yandong17} of the minority classes to the majority. Zhang~\etal~\cite{Zhang_8237840} proposed a range loss to minimize the intra-class variance based on the largest intra-class distances (ranges) computed regardless of the imbalanced class size. Our method complements these methods by providing a data structure-aware loss function that enforces margins between local data to reduce the imbalance in any local neighborhood.

\noindent
{\bf Deep face attribute prediction.} Face attributes are useful as mid-level features for many applications like face verification~\cite{berg13,kumar2011describable}. It is challenging to predict them from unconstrained face images due to the large facial variations. Most existing methods utilize part-based models to extract features from the localized part regions, and then train SVM classifiers to predict the presence of an array of face attributes,~\eg,~``male'' and ``smile''. For example, Kumar~\etal~\cite{kumar2011describable} extracted HOG-like features from various local face regions for attribute prediction. Recent deep learning methods~\cite{liu15,Ning14,Kalayeh17} excel by learning powerful features. Kalayeh~\etal~\cite{Kalayeh17} further combined a deep semantic segmentation network to guide attribute prediction to the corresponding local region. These studies, however, share a common drawback: they neglect the class imbalance issue in those relatively rare attributes like ``big nose'' and ``bald''. To our knowledge, only two works handle class imbalance in attribute prediction. The Mixed Objective Optimization Network (MOON)~\cite{Rudd2016} re-weights attributes in a cost-sensitive manner, and the Class Rectification Loss (CRL)~\cite{Dong_2017_ICCV} performs online regularization for minority attribute classes in batch. Our method shows a stronger imbalanced learning ability with a new loss function.

\section{Learning Deep Representation from Class-Imbalanced Data}
\label{sec:method}

Given an imagery dataset with imbalanced class distribution, our goal is to learn an embedding function $f(x)$ from an image $x$ into a feature space $\mathbb{R}^d$, such that the embedded features are discriminative with local class imbalance ameliorated. We constrain this embedding to live on a $d$-dimensional hypersphere, \ie,~$||f(x)||_2 = 1$. Such normalization is commonplace in existing embedding methods (\eg,~triplet embedding~\cite{Schroff15}), in order to achieve scale invariance under different image conditions,~\eg,~lighting, contrast and so on.

To achieve the above learning goal, we start by giving a brief review of the challenges with existing embedding methods that hinder their performance on class-imbalanced data. They will motivate our work to follow.

\subsection{Challenges with Existing Embedding Methods}

\noindent
{\bf Triplet loss.} The triplet loss~\cite{Schroff15} and contrastive loss~\cite{Yi14} are two popular approaches to learn a Euclidean embedding function $f(x)$. Some triplet variants~\cite{NIPS6200,NIPS6368} were recently proposed for improvements. For simplicity here, we only use the vanilla triplet loss to exemplify the inefficacy of this line of methods when handling class imbalance.

Triplet embedding is trained on a set of triplets $\mathcal{P}=\left\{ (x_i,x_i^p,x_i^n) \right\}$ where $x_i$ is the anchor point, associated with the positive and negative examples $x_i^p$ and $x_i^n$. The sampling of triplets is usually semantic, informed by class labels: $(x_i,x_i^p)$ come from the same class, and $(x_i,x_i^n)$ come from different classes. Then the goal is to enforce semantic similarity between the feature embeddings $f(\cdot;\Theta)$ extracted by one CNN with parameters $\Theta$ (we will omit $\Theta$ later for brevity). More precisely, the objective is to push away the negative example $x_i^n$ from the anchor $x_i$ in the embedding space by a Euclidean distance margin $g>0$ compared to the positive example $x_i^p$:
\begin{equation}
\label{eq1}
D(f(x_i),f(x_i^p))+g<D(f(x_i),f(x_i^n)),
\end{equation}
where $D(f(x_i),f(x_j))=\|f(x_i)-f(x_j)\|_2^2$ is the Euclidean distance.

The cost function is defined as:
\begin{equation}
\label{eq2}
J_{tri} = \frac{1}{|\mathcal{P}|} \sum_{i\in \mathcal{P}} \left[ D(f(x_i),f(x_i^p)) - D(f(x_i),f(x_i^n))+g \right]_+,
\end{equation}
where $[\cdot]_+=max(0,\cdot)$ denotes the hinge function.

\begin{figure*}[t]
\begin{center}
\includegraphics[width=1.0\linewidth]{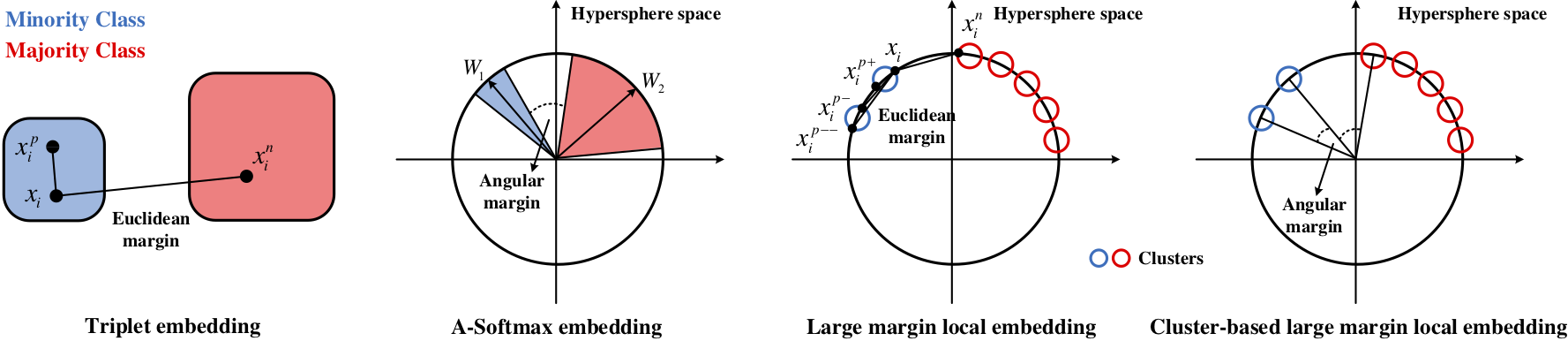}
\end{center}
\vspace{-1.5em}
\caption{The 2-D feature space of triplet loss~\cite{Schroff15}, A-Softmax loss~\cite{liu2017sphereface}, Large Margin Local Embedding (LMLE)~\cite{huang2016lmle} and the proposed Cluster-based Large Margin Local Embedding (CLMLE). Class imbalance is exemplified in a binary-class case. The triplet and A-Softmax losses enforce Euclidean and angular margins respectively at class-level, assuming each class can be captured by a single mode. Such unimodal discrimination imposes too strong of a requirement, and may fail to collapse the majority class with larger variation and lead to class overlap. The LMLE enforces Euclidean margins among quintuple examples sampled from the intra- and inter-class local clusters. Such constraint preserves discrimination in local neighborhood and helps form local class boundaries that are insensitive to the imbalanced class size. The proposed CLMLE samples the entire cluster distributions instead to address the inefficiency and inconsistency issues with quintuplet sampling in LMLE. CLMLE also naturally facilitates the derivation of angular margins between cluster distributions on the unit hypersphere.}
\label{fig2}
\end{figure*}

It is widely observed that the convergence rate and performance of triplet embeddings are hindered by two factors: 1) the cubic growth of the number of triplets, and 2) the penalty on individual triplets not being necessarily consistent. Hard negative mining~\cite{Schroff15} (semi-hard negative mining in the paper) is one of the ways to improve the triplet quality and hence the learning efficiency. However, hard mining still cannot conquer the limitation of the triplet loss in capturing the structure of imbalanced data. Specifically, the similarity information is only extracted at the \textit{class-level} by triplets. This tends to equally collapse the intra-class variation and encourage unimodal discrimination. But it is easy to tell that the difficulty of collapsing different classes varies. Fig.~\ref{fig2} shows an imbalanced binary case. Obviously it is more difficult to collapse the majority class with larger variation than minority class. In case of collapsing failure, it would immediately lead to the invasion of imposter neighbors or even domination of the majority class in local neighborhood.

\noindent
{\bf A-Softmax loss.} The A-Softmax loss in SphereFace~\cite{liu2017sphereface} enhances the discrimination power of Softmax by imposing the angular margin in a hypersphere space. However, it still suffers from the unimodal assumption of class distributions, which is insufficient on imbalanced data (see Fig.~\ref{fig2}). In the traditional Softmax loss, the class score $s_j$ for sample $x_i$ can be written as $s_j=\pmb{W}_j^T f(x_i) = \| \pmb{W}_j\|\| f(x_i)\| \cos(\theta_j)$ where $\pmb{W}_j$ is the $j$-th column of fully connected layer $\pmb{W}$, and $\theta_j$ is the angle between $\pmb{W}_j$ and $f(x_i)$. In A-Softmax loss, each $\pmb{W}_j$ is normalized $\|\pmb{W}_j\|=1,\forall j$ and the loss becomes:
\begin{equation}
\label{eq3}
J_{ang} = \frac{-1}{|\mathcal{P}|} \sum_{i\in \mathcal{P}} \log \left( \frac{e^{\| f(x_i)\|\psi(\theta_{y_i})}}{e^{\| f(x_i)\|\psi(\theta_{y_i})}+\sum_{j\ne y_i} e^{\| f(x_i)\|\cos(\theta_j)}}\right),
\end{equation}
where $x_i$ has the class label $y_i \in [1,C]$, and $j\ne y_i$ are the other labels in the training set $\mathcal{P}$. The function $\psi(\cdot)$ incorporates an angular margin for the class $y_i$ on unit hypersphere. Obviously, due to the enforcement of angular margins at class level, the intra-class structures are lost again and the issues with triplet loss on imbalanced data apply to the A-Softmax loss as well.

\noindent
{\bf Large Margin Local Embedding (LMLE).} Our previously proposed LMLE~\cite{huang2016lmle} addresses the class-imbalance issue by assuming that each class's distribution can be represented as a variant number of clusters. Then we are able to draw balanced class boundaries only among the involved local clusters, not at the whole class level anymore. To this end, quintuplet instances (see Fig.~\ref{fig2}), sampled from clusters both within and between classes, are used as hard examples to form implicit local boundaries:

\small
\begin{packed_itemize}
\item $x_i\;\;\;\;\,\,$: an anchor,
\item $x_i^{p+}\;\,\,$: the anchor's most distant within-cluster neighbor,
\item $x_i^{p-}\;\,\,$:  the nearest within-class neighbor of the anchor, but from a different cluster,
\item $x_i^{p--}$: the most distant within-class neighbor of the anchor,
\item $x_i^{n}\;\,\,\,\,$: the nearest between-class neighbor of the anchor.
\end{packed_itemize}
\normalsize

We wish to ensure that the following relationship holds in the embedding space:
\begin{eqnarray}
\label{eq4}
   &&D(f(x_i),f(x_i^{p+}))<D(f(x_i),f(x_i^{p-}))\nonumber\\
< \!\!\!\!\!\!\!\!\!\! &&D(f(x_i),f(x_i^{p--}))<D(f(x_i),f(x_i^{n})).
\end{eqnarray}

Such a fine-grained relationship embraces the multi-modality of class distribution: it preserves not only locality across the same-class clusters but also discrimination between classes. As a result, it is capable of preserving discrimination in any local neighborhood, and forming local class boundaries with the most discriminative samples. Other irrelevant samples in a class are effectively ``ignored'' for class separation, making the local boundaries insensitive to imbalanced class sizes. To enforce the relationship in Eq.~(\ref{eq4}), a \textit{triple-header hinge loss} is formulated to constrain three margins between the four Euclidean distances:
\begin{eqnarray}
\label{eq5}
  \!\!\!\!\!\!\!\!\!\!\!\!\!\! && J_{lmle}= \frac{1}{|\mathcal{P}|} \sum_{i \in \mathcal{P}} (\varepsilon_i+\tau_i+\sigma_i), \;\;\; s.t.: \nonumber\\
  \!\!\!\!\!\!\!\!\!\!\!\!\!\! && \left[ g_1+D(f(x_i),f(x_i^{p+}))-D(f(x_i),f(x_i^{p-}))\right]_+\leq \varepsilon_i, \nonumber\\
  \!\!\!\!\!\!\!\!\!\!\!\!\!\! && \left[ g_2+D(f(x_i),f(x_i^{p-}))-D(f(x_i),f(x_i^{p--}))\right]_+\leq \tau_i, \nonumber\\
  \!\!\!\!\!\!\!\!\!\!\!\!\!\! && \left[ g_3+D(f(x_i),f(x_i^{p--}))-D(f(x_i),f(x_i^{n}))\right]_+\leq \sigma_i, \nonumber\\
  \!\!\!\!\!\!\!\!\!\!\!\!\!\! && \forall i, \; \varepsilon_i \geq 0, \; \tau_i \geq 0, \; \sigma_i \geq 0
\end{eqnarray}
where $\varepsilon_i,\tau_i,\sigma_i$ are the slack variables, $g_1,g_2,g_3$ are the enforced Euclidean distance margins that can be explicitly determined on the hypersphere manifold.

Despite being effective on imbalanced data, quintuplets have similar sampling issues as triplets --- the exponential growth of quintuplet number and the potential inconsistency between sampled quintuplets, both of which can hinder the convergence rate and quality. This work generalizes to sample the entire cluster distributions for learning. We will demonstrate the significantly improved learning efficiency and consistency, and hence better discrimination in the imbalanced context.

\subsection{Cluster-based Large Margin Local Embedding}

We propose to learn Cluster-based Large Margin Local Embedding (CLMLE) from all the examples in contextual clusters, rather than only hard examples (quintuplets) in clusters. Clustering techniques are employed to capture the local distributions of clusters for every class. Since our feature vectors $\{f(x_i)\}$ are always normalized onto the unit hypersphere (see Fig.~\ref{fig2}), we choose the spherical \textit{k}-means algorithm for clustering. Suppose we have the training set $\mathcal{P}=\left\{(x_i,y_i)\right\}_{i=1}^L$ with sample $x_i$ and class label $y_i \in [1,C]$. Then for each class $c$, we have $K$ cluster assignments:
\begin{eqnarray}
\label{eq6}
\mathcal{I}_1^c,\dots,\mathcal{I}_K^c &=& \argmax_{\mathcal{I}_1^c,\dots,\mathcal{I}_K^c} \sum_{k=1}^K \sum_{i\in \mathcal{I}_k^c} \trans{f(x_i)} \mu_k^c, \\
\mu_k^c &=& \frac{1}{|\mathcal{I}_k^c|} \sum_{i\in \mathcal{I}_k^c} f(x_i), \;\; \mu_k^c = \frac{\mu_k^c}{\| \mu_k^c\|},
\end{eqnarray}
where $\{\mu_k^c\}_{k=1}^K$ are the cluster centroids in class $c$. We generate clusters with the same size $|\mathcal{I}_k^c|=l$ to ensure an equal complexity to collapse these clusters during embedding learning, and to draw implicit balanced boundaries between them. The number of clusters $K = \lfloor L_c/l \rfloor$ is adaptively determined for each class that has size $L_c$. Note here we use the inner product as the similarity metric, which is more suitable than Euclidean distance on the unit hypersphere. It also facilitates the derivation of our angular margins, as will be elaborated next.

During training, we are interested in a global characterization of data neighborhoods and penalizing any overlap of cluster distributions in a coherent way. Specifically, at each training iteration, we retrieve for a query cluster $\mathcal{I}_1$ its nearest clusters $\{\mathcal{I}_m\}_{m=1}^{M-1}$ from both the same and different classes in the neighborhood. We define $\mu_m$ as the centroid of cluster $\mathcal{I}_m$, and let $c(\mu_m)$ denote the class label of $\mu_m$. Then local discrimination can be achieved by making all the clusters inter-distinct and intra-compact, so that each cluster is able to represent a unique mode for future discrimination. Concretely, we maximize the intra-cluster similarity between $f(x_i)$ and its containing cluster centroid $\mu_m$, while minimizing the inter-cluster similarity between $f(x_i)$ and other cluster centroids $\{\mu_{k:k\ne m}\}$. We use the inner product again to characterize similarity and define our loss function as follows:
\begin{equation}
\label{eq8}
J_{clmle} = \frac{1}{Ml} \sum_{m=1}^M \sum_{i \in \mathcal{I}_m} \left[ -\log \frac{e^{\trans{f(x_i)} \mu_m - a}} {\sum_{k: k\ne m} e^{\trans{f(x_i)} \mu_k}} \right]_+,
\end{equation}
where $a$ is an angular margin on the unit circle, the desired angular gap between the cluster centroid $\mu_m$ and others.

The above loss function actually penalizes the probability assigned to the example $x_i$ of a particular cluster $\mu_m$ under the distribution of another by a large margin $a$. This will gradually collapse each cluster distribution into a small region and form safe margins between adjacent regions on the hypersphere. More importantly, those ``marginal'' clusters near the class bounds will implicitly draw a balanced classification boundary between them. As a result, the class imbalance is effectively reduced locally, ignoring the impact of other far-away clusters. To guarantee sufficient class separation between those marginal clusters, we inject the information of class label into Eq.~(\ref{eq8}), enforcing a larger angular margin for those between-class clusters than for same-class clusters:
\begin{eqnarray}
\label{eq9}
 J_{clmle} = \frac{1}{Ml} \sum_{m=1}^M \sum_{i \in \mathcal{I}_m} && \!\!\!\!\!\!\!\!\!\! \left[ -\log \frac{e^{\trans{f(x_i)} \mu_m - a_1}} {\sum_{k:c(\mu_k) \ne c(\mu_m)} e^{\trans{f(x_i)} \mu_k}} \right]_+ \nonumber\\
&& \!\!\!\!\!\!\!\!\!\!\!\!\!\!\!\!\!\!\!\!\!\!\!\!\!\!\!\!\!\!\!\!\!\!\!\!\!\!\!\! + \left[ -\log \frac{e^{\trans{f(x_i)} \mu_m - a_2}} {\sum_{k: k\ne m, c(\mu_k) = c(\mu_m)} e^{\trans{f(x_i)} \mu_k}} \right]_+,
\end{eqnarray}
where $a_1$ and $a_2$ are the inter-cluster angular margins enforced between different classes and within the same class, respectively. We will introduce how to explicitly derive $a_1$ and $a_2$ later.

\noindent
{\bf Comparisons with related embedding methods.} The new loss in Eq.~(\ref{eq9}) is similar in spirit to the probabilistic model of Nearest Class Mean (NCM)~\cite{Mensink2013}. The main difference is that we adopt a non-uniform assumption for the class distribution, each modeled as a mixture of homogeneous clusters. This is well suited to class-imbalanced data and allows for local discrimination using only a balanced subset of clusters. While in NCM, the homogeneous class assumption is adopted, which is invalid in imbalanced settings. Another difference is that we learn deep feature representations rather than use fixed ones as in NCM.

\begin{figure}[t]
\begin{center}
\includegraphics[width=1.0\linewidth]{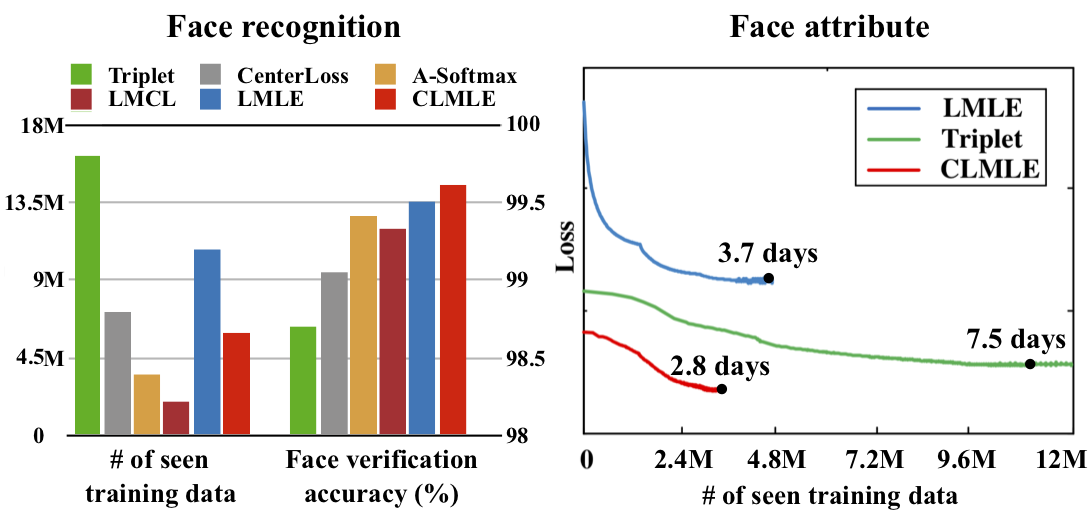}
\end{center}
\vspace{-1.5em}
\caption{Speed and performance analyses in face recognition (on LFW~\cite{LFWTech}) and face attribute (on CelebA~\cite{liu15}) tasks. We compare convergence speed by the number of seen training data, which is fair and independent of differences in GPU, batch size used and other factors. We also report training times for the face attribute task. The compared methods are triplet loss~\cite{Schroff15}, Center loss~\cite{wen2016}, A-Softmax~\cite{liu2017sphereface}, LMCL~\cite{2018Wang}, Large Margin Local Embedding (LMLE)~\cite{huang2016lmle} and our Cluster-based Large Margin Local Embedding (CLMLE). CLMLE strikes a good performance-speed trade-off.}
\vspace{-0.8em}
\label{fig3}
\end{figure}

When compared with the embedding methods based on individual triplets~\cite{Schroff15} or quintuplets~\cite{huang2016lmle}, our method has two main merits: 1) Our sampling of entire clusters is linear to the size of training data $\mathcal{O}(N)$, thus significantly avoids the sampling complexity of triplets $\mathcal{O}(N^3)$ and quintuplets $\mathcal{O}(N^5)$ and improves the training efficiency. 2) Our loss informed of entire cluster distributions has a sufficient insight of contextual neighborhoods. By separating all clusters at once, the training is more coherent and globally consistent than training with individual triple or quintuple examples. Fig.~\ref{fig3} demonstrates our CLMLE indeed converges much faster than the triplet loss~\cite{Schroff15} and quintuplet-based LMLE~\cite{huang2016lmle}, while achieving better performance at the same time. In comparison to other state-of-the-art embedding methods (\eg,~Center loss~\cite{wen2016}), CLMLE strikes a better performance-speed trade-off.

The recent Class Rectification Loss (CRL)~\cite{Dong_2017_ICCV} performs regularization for the ``hard'' minority classes in every batch. By contrast, our CLMLE method performs equal learning for both minority and majority classes over batches. More importantly, the clustering step for all classes informs our learning process of the global feature distribution. It encourages faster and better learning convergence than the purely online method CRL, at negligible cost. On the CelebA dataset, for instance, CLMLE converges about 15 times faster than CRL.

\noindent
{\bf Angular margin derivation.} One good characteristic of our loss function in Eq.~(\ref{eq9}) is that the margins $a_1$ and $a_2$ can be explicitly derived following a geometric intuition. These margins are designed angular, which translates well to the inner product based similarity metric on a unit hypersphere. Angular margins also share the favorable properties of scale and rotation invariance.

\begin{figure}[t]
\begin{center}
\includegraphics[width=1.0\linewidth]{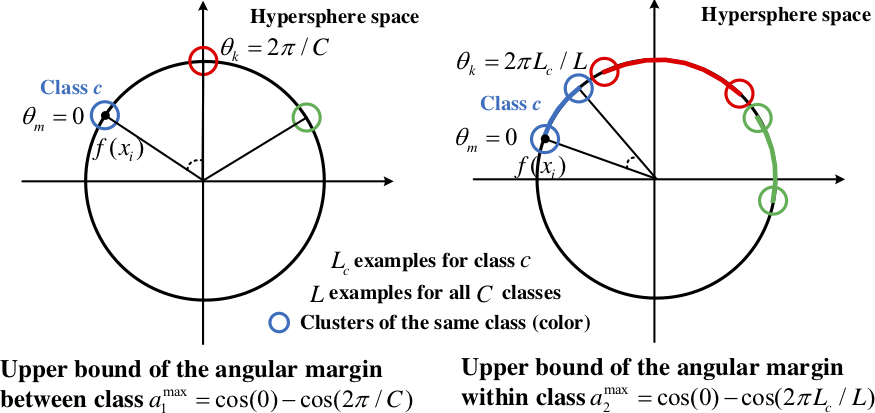}
\end{center}
\vspace{-1.5em}
\caption{Extreme 2D cases for deriving the upper bounds of inter-cluster angular margins between class ($a_1^{max}$) and within class ($a_2^{max}$).}
\vspace{-0.8em}
\label{fig4}
\end{figure}

Fig.~\ref{fig4} illustrates how to set $a_1$ and $a_2$ properly. Obviously, their lower bounds are zero. Then we derive their theoretical upper bounds $a_1^{max}$ and $a_2^{max}$ in 2D space, to provide a reasonable parameter range for grid search during training. Recall that $a_1$ denotes the inter-cluster angular margin from different classes, and $a_2$ denotes the one within the same class. It is easy to first find the angular gap would peak when all clusters or even the whole classes are collapsed into single points on the hypersphere and they are all widely separated in between. In such extreme case, each unnormalized cluster centroid $\mu_m=\sum_{i \in \mathcal{I}_m}f(x_i)/l$ is identical to the cluster members thus has unit norm too. The inner product $f(x_i)^T \mu_k=\| f(x_i)\|\| \mu_k\| \cos(\theta_k)=\cos(\theta_k)$ in Eq.~(\ref{eq9}) now depends on the angle $\theta_k$ only, while the intra-cluster similarity is fixed as $f(x_i)^T \mu_m=\cos(\theta_m)=\cos(0)$. Hence we can define the maximum angular margin in the form of $\cos(0)-\cos(\theta_k)$. Easily, we have $a_1^{max}=\cos(0)-\cos(2\pi /C)$ when each class becomes a single point with the maximum inter-class angle $\theta_k=2\pi /C$. $a_2^{max}=\cos(0)-\cos(2\pi L_c/L)$ when the considered clusters are most far apart in their belonging class that occupies a proportion $L_c/L$ of the hyperspherical space.

\noindent
{\bf Cluster updates.} It is worth noting that the cluster assignment $\mathcal{I}_1^c,\dots,\mathcal{I}_K^c$ for each class $c$ in Eq.~(\ref{eq6}) is initialized using the features from a pre-trained CNN. Since the feature representations $\{f(x_i)\}$ are updated continuously during training, we should gradually update the clusters as well on the newly learned features to reflect their true distribution. To this end, for each class, we maintain a running index of clusters and refresh it after a fixed number of iterations using the latest features. The computational cost of the clustering algorithm of spherical \textit{k}-means is negligible compared to the cost of learning CNN features.

\begin{figure*}[t]
\begin{center}
\includegraphics[width=1.0\linewidth]{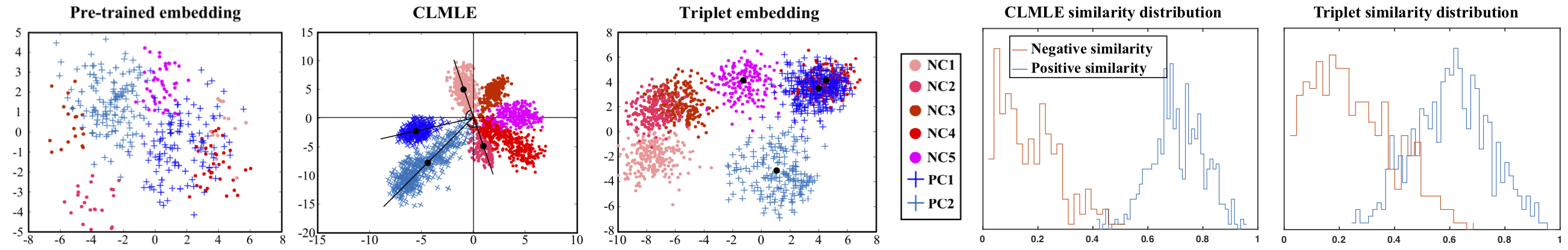}
\end{center}
\vspace{-1.5em}
\caption{The 2-D feature space using t-SNE~\cite{tsne} and pairwise feature similarity for one binary face attribute from the CelebA dataset~\cite{liu15}. We only show 2 Positive Clusters (PC) and 5 Negative Clusters (NC) to represent the class imbalance. The embedding of a pre-trained model, our CLMLE, and triplet embedding are compared. We can see that between-class clusters (with different colors) are well separated in CLMLE, but they are overlapped in triplet embedding, leading to overlapping binary score distributions.}
\label{rebuttal_fig5}
\end{figure*}

\noindent
{\bf Embedding visualization.} Fig.~\ref{rebuttal_fig5} visualizes the 2-D space of the initial embedding and final converged CLMLE in an imbalanced face attribute example. Triplet embedding~\cite{Schroff15} as a representative unimodal learning method, is included for comparison. Since triplets operate at the class-level and homogeneously collapse each class, we see their inability to capture the fine-grained variations (represented by clusters) within the imbalanced classes, which leads to a large overlap between the minority (positive) and majority (negative) classes. Such issue equally applies to the other unimodal methods of the contrastive loss~\cite{Yi14}, advanced triplet loss~\cite{NIPS6200}, and angular losses like A-Softmax~\cite{liu2017sphereface} and L-Softmax~\cite{Liu2016}. By contrast, our CLMLE explores the multimodal property of class by enforcing angular margins both within and between classes. As demonstrated in the figure, this helps to learn unique clusters of balanced size in each class, which are able to draw balanced local class boundaries for discrimination. We will show this can also boost the feature generalization in open-set scenarios (for face verification) that often require discriminative features with built-in margins. It is worth noting that Wang~\etal~\cite{Jiang14} aim to learn fine-grained similarity within class as well, but they do not explicitly rely on clustering techniques to model the within-class variations.

\subsection{Overall Training Procedure with Re-Sampling and Cost-Sensitivity}

During training, we construct one mini-batch with a query cluster $\mathcal{I}_1$ and its retrieved $M-1$ nearest clusters $\{\mathcal{I}_m\}_{m=1}^{M-1}$ by computing $\mu_1^T \mu_m$. We greedily make sure the retrieved clusters would come from both the same and different classes. Since the cluster size $l=200$ is not small, the batch size $Ml$ will be very large even if we only retrieve a small number of clusters $M$. This is not viable for training due to memory constraints. On the other hand, we observed empirically that using only a few clusters (\ie,~low cluster diversity) hurts performance. Thus we choose to sample a small portion of data in each considered cluster to increase the cluster number $M$ while maintaining a reasonable batch size. In practice, we randomly sample 20 examples out of $l=200$ from each of $M=12$ clusters. The batch size is 240. The cluster centroid in Eq.~(\ref{eq9}) is then approximated as $\hat{\mu}_m=\sum_{i=1,\dots,20, i \in \mathcal{I}_m}f(x_i)/20$. Such cluster data sampling is simply repeated during CNN training. It avoids large information loss as in traditional random under-sampling techniques. When compared with over-sampling, it introduces no artificial noise.


So far, one important problem is still left unaddressed: how to sample the query cluster $\mathcal{I}_1$ in mini-batch. We found a random sampling strategy is not ideal for performance. Here we simply follow the common practices of re-sampling and cost-sensitive techniques as detailed below. Section~\ref{sec:result} will quantify their efficacy systematically.

\noindent
{\bf Re-sampling of query cluster $\mathcal{I}_1$.} To ensure adequate learning for all classes, we sample $\mathcal{I}_1$ evenly from both majority and minority classes. To determine the exact query cluster to use from the chosen class, we borrow the idea of ~\cite{BuloNK17} to pick $\mathcal{I}_1$ as the one with the highest observed loss in class. This allows us to adapt to the current feature distribution and focus on the hardest cluster that has large overlap with neighboring ones. In practice, the loss of a cluster is computed by averaging the losses of cluster members that are cached online.

\noindent
{\bf Cost-sensitive learning in batch.} Note the query and retrieved clusters are most likely to be class-imbalanced in one mini-batch. We follow the commonly used cost-sensitive approach~\cite{CaesarUF15,MostajabiYS15} to scale the loss of each sample by the inverse class frequency in mini-batch. For multi-way classification (\eg,~in face identification), this effectively gives those less frequent classes more importance. For the problem of predicting multiple attribute labels from a face, cost-sensitivity helps even more by being able to re-balance all labels simultaneously. Note re-sampling multi-label data is structurally infeasible because sampling to balance one label will affect the sampling of others.

\vspace{0.5em}

\noindent
{\bf Overall training procedure for CLMLE:}
\vspace{-0.3em}
\begin{packed_enumerate}
  \item Cluster for each class by spherical \textit{k}-means using the latest features. Initially, we use the features extracted by the pre-trained CNN.
  \item For CNN training, repeatedly construct mini-batches with one query cluster (with highest loss) from a random class, and the $M-1=11$ nearest clusters retrieved from both the same and different classes.
  \item Randomly sample 20 examples for each of the $M=12$ clusters in batch, and compute their cluster centroids.
  \item Compute the loss in Eq.~(\ref{eq9}) with cost-sensitivities (by inverse class frequency). Back-propagate the gradients to update the CNN parameters and feature embeddings.
  \item Alternate between step 1 and 2-4 periodically until convergence (often within 5 alternation rounds).
\end{packed_enumerate}

\section{Fast Evaluation with Nearest Clusters}

Our learned CLMLE offers crucial feature representations for accurate evaluation on imbalanced data. We choose a simple \textit{k}-nearest cluster algorithm which is consistent with our training objective. The nearest neighbor rule is appealing due to its non-parametric nature, and it is easy to extend to new classes without retraining.

Specifically, for a query $q$, we retrieve its $N$ nearest clusters $\{\mathcal{I}_m\}_{m=1}^{N}$ from all the training classes by computing $f(q)^T \mu_m$, and decide its class label by:
\begin{equation}
\label{eq10}
y_q= \underset{c=1,\ldots,C}{\arg\max} \frac{\underset{m:c(\mu_m)=c}{\min} e^{\trans{f(q)} \mu_m}} {\sum_{k:c(\mu_k) \ne c} e^{\trans{f(q)} \mu_k}},
\end{equation}
where the label $y_q$ is predicted as the class whose least similar cluster is more similar than the clusters from other classes by the largest margin.

Such testing procedure makes the nearest neighbor search a function of the cluster number $\lfloor L/l \rfloor$ rather than of the example number $L$. We further speed up the cluster-wise search using the KD-tree~\cite{Silpa08} whose runtime is logarithmic in the cluster number ($\lfloor L/l \rfloor$) with a complexity of $\BigO(L/l\log(L/l))$. This leads to up to three orders of magnitude speedup over standard example-wise search in practice, making it easy to scale to large datasets. It is worth mentioning that we use such a simple testing algorithm to show the efficacy of our learned features. Better performance is expected with the use of more elaborated algorithms.

\section{Experiments}
\label{sec:result}

\begin{figure}[t]
\begin{center}
\includegraphics[width=1.0\linewidth]{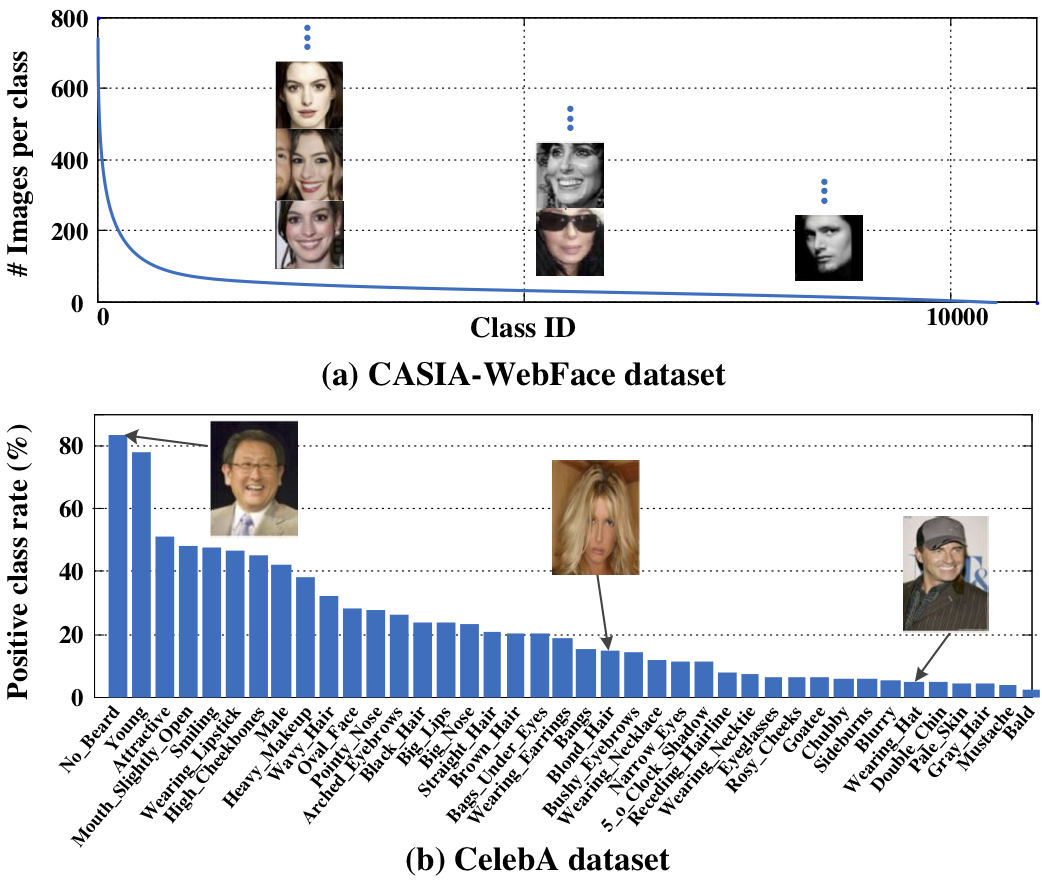}
\end{center}
\vspace{-1.5em}
\caption{Imbalanced data distribution for face recognition and face attribute prediction. (a) Long-tailed distribution of image number per class on CASIA-WebFace dataset~\cite{YiLLL14a}. (b) 40 binary face attributes on CelebA dataset~\cite{liu15}, each with imbalanced positive and negative samples.}
\label{fig6}
\end{figure}

We study the face recognition and face attribute prediction tasks, both with large-scale imbalanced datasets (see Fig.~\ref{fig6}). The face recognition task is cast as either an identification (multi-way classification of a face) or verification (binary classification of a face pair) problem, evaluated in the open-set scenario. As shown in the figure, the available training dataset of CASIA-WebFace~\cite{YiLLL14a} is highly imbalanced. Some classes of the 10k subjects have hundreds of images, while about 39\% of them have no more than 20 images. The average number of images per class is 42.8. Such data imbalance makes it difficult for learning equally good class representations which are essential for comparing paired images from different classes. The face attribute prediction task is cast as a (closed-set) multi-task classification problem. We aim to predict 40 binary attributes simultaneously, each with imbalanced positive and negative samples (\eg,~for ``Bald'' attribute: 2\% vs. 98\%). In this case, the variant imbalance level in multi-label data makes the problem even harder.

\noindent
{\bf Parameters.} Our CNN is trained using \texttt{Caffe} with fixed momentum 0.9, weight decay $\lambda=0.0005$. The learning rate starts with 0.1 and 0.001 for face recognition and attribute prediction respectively, and is divided by 10 when the performance plateaus. We have cluster size $l=200$ and $M=12$ clusters in one batch. The down-sampled $M$ clusters leads to the batch size 240. For testing, the optimal number of retrieved clusters $N$ in Eq.~(\ref{eq10}) is searched from $20:10:200$ on validation set. The task-specific network architecture and prior features for initial clustering are summarized in Table~\ref{tb1}. Note the prior features are not critical to the final results because we will gradually update the deep features in alternation with the clustering process. The alternation happens every 2k - 5k iterations, depending on the task-specific convergence rate. Usually different prior features converge to similar results. For our face verification (on LFW~\cite{LFWTech}) and attribute prediction experiments, different prior features lead to accuracy difference of no more than 0.02\% (pre-trained by softmax vs. triplet) and 0.3\% (pre-trained by face recognition vs. multi-attribute softmax classification) respectively.

\noindent
{\bf Computational cost.} For face recognition and face attribute prediction, it takes about 1.5 and 3 days respectively to train CLMLE on GPU (NVIDIA Tesla K40), both converged within 5 alternation rounds. The clustering process in each round incurs negligible cost compared to feature learning. CLMLE has about 1.3 - 2 times faster convergence than its earlier version LMLE~\cite{huang2016lmle} (see Fig.~\ref{fig3}) thanks to the avoidance of quintuplet-based sampling and learning. For testing, it takes 10ms to extract features, and the cluster-wise kNN search is typically 1000 times faster than standard kNN. This enables real-time application to large-scale problems with hundreds of thousands to millions of samples.

\begin{table}[t]
\caption{The CNN architecture and prior features for initial clustering in our considered imbalanced tasks.}
\centering
\resizebox{0.95\linewidth}{!}{
\begin{tabular}{ c|c|c }
\hline
Task & Network & Prior features \\
\hline
\hline
Face recognition & Same w.r.t.~\cite{liu2017sphereface} & Pre-trained by softmax \\

Face attributes & Same w.r.t.~\cite{Yi14} & DeepID2 features in~\cite{Yi14} \\
\hline
\end{tabular}
}
\label{tb1}
\end{table}

\subsection{Experimental Settings}

\noindent
{\bf Face recognition.} All faces and landmarks are detected by MTCNN~\cite{Zhang_7553523} for the training and testing images. The detected landmarks (two eyes, nose and mouth corners) are used for similarity transformation, and the faces are cropped to $112\times 96$ pixels. Each pixel (in $[0,255]$) in cropped images is normalized by subtracting 127.5 and then dividing by 128.

\underline{For training}, we use the same CNN architecture as in~\cite{liu2017sphereface,2018Wang} with 64 convolutional layers based on residual units. This enables fair comparison with the recent strong methods and their variants. For the same reason, we use a small training set, the publicly available CASIA-WebFace dataset~\cite{YiLLL14a} which contains 0.45M face images from 10,575 identities. The training images with identities appearing in our test sets are already removed. Note the scale of our training data is much smaller than that of other private datasets used in DeepFace~\cite{Taigman14} (4M), VGGFace~\cite{Parkhi15} (2M) and FaceNet~\cite{Schroff15} (200M). In practice, we apply data augmentation by flipping the training images horizontally.

\underline{For testing}, we extract the deep features $f(x)$ from the output of the FC1 layer. The features of the original image and flipped image are concatenated to obtain the final face representation. The inner product between normalized features is then calculated as the similarity score. In the closed-set scenario with fixed class set, face recognition can be done by using the \textit{k}-nearest cluster rule in Eq.~(\ref{eq10}); while in open-set scenarios with new classes, identification and verification are respectively conducted by ranking and thresholding the similarity scores following the convention, which will be shown to still benefit from our learned CLMLE.

We test on three popular large-scale datasets: LFW~\cite{LFWTech}, YTF~\cite{Wolf_2011} and MegaFace~\cite{kemelmacher2016megaface}. LFW dataset contains 13,233 web images from 5,749 face identities captured in unconstrained conditions. YTF dataset contains 3,425 videos from 1,595 identities. Each video varies from 48 to 6,070 frames, with an average length as 181.3 frames. Both datasets exhibit large facial variations in pose, expression and lighting. We follow the standard protocol of unrestricted with labeled outside data~\cite{LFWTech} for both datasets, and test on 6k face pairs from LFW and 5k video pairs from YTF. MegaFace dataset is a very challenging benchmark for face recognition at the million scale of distractors. The gallery set in MegaFace contains more than 1M images from 690k identities, while the probe set consists of two existing datasets --- Facescrub and FGNET. We choose the larger Facescrub dataset as our probe set which contains 106,863 face images from 530 celebrities. We report the face identification and verification results under two protocols (small or large training set).

\underline{For evaluation}, we not only use the simple metrics of the identification or verification accuracy, but also the True Accept Rate (TAR) at fixed False Accept Rate (FAR) as well as ROC curves that can take into account any imbalance in testing pairs.

\noindent
{\bf Face attributes.} We use the CelebA dataset~\cite{liu15} that contains 202,599 images from 10,177 identities, each with about 20 images. Every face image is annotated with 40 attributes and 5 key points to align the image to $55\times47$ pixels. We partition the dataset following~\cite{liu15}: the first 162,770 images (\ie,~8k identities) for training (10k images for validation), the following 19,867 images for training SVM classifiers for the PANDA~\cite{Ning14} and ANet~\cite{liu15} methods, and the remaining 19,962 images for testing. The identities are non-overlapping in these splits. During training, horizontal flipping is applied for data augmentation. We use the CNN architecture from~\cite{Yi14} as in ANet~\cite{liu15}, LMLE~\cite{huang2016lmle} and CRL~\cite{Dong_2017_ICCV}. One extra 64-d FC layer is learned for each binary attribute via Eq.~(\ref{eq9}) in a multi-task manner.

For testing, we extract the deep features $f(x)$ for each attribute from its FC layer, and classify attributes via Eq.~(\ref{eq10}) under the closed-set classification scenario. To account for the imbalanced positive and negative samples for each attribute, we adopt a balanced accuracy metric $accuracy = 0.5(t_p/N_p + t_n/N_n)$, where $N_p$ and $N_n$ are the numbers of positive and negative samples, while $t_p$ and $t_n$ are the numbers of true positives and true negatives. Note this metric differs from the one employed in~\cite{liu15},~\ie,~$accuracy = ((t_p + t_n) / (N_p + N_n))$ which can be biased to the majority class.

\subsection{Face Recognition}

\begin{table}[t]
\caption{Face verification accuracy (\%) on LFW~\cite{LFWTech} and YTF~\cite{Wolf_2011} datasets. Most methods in the first cell use large-scale outside data that are not publicly available (+ denotes data expansion). The second cell includes recent imbalanced learning methods. The state-of-the-art loss functions in the second-to-last cell and ours in the last cell use the small training data (CASIA-WebFace~\cite{YiLLL14a}: 0.45M) and the same 64-layer CNN model for fair comparison.}
\centering
\resizebox{1.0\linewidth}{!}{
\begin{tabular}{ c|c|c|c|c }
\hline
Method & \#Nets & Train data & LFW & YTF \\
\hline \hline
DeepFace~\cite{Taigman14} & 3 & 4M & 97.35 & 91.4 \\
FaceNet~\cite{Schroff15} & 1 & 200M & 99.63 & 95.1 \\
Web-scale~\cite{TaigmanYRW15} & 4 & 4.5M & 98.37 & - \\
VGG Face~\cite{Parkhi15} & 1 & 2.6M & 98.95 & \textbf{97.3} \\
DeepID2+~\cite{SunWT15} & 25 & 0.3M & 99.47 & 93.2 \\
Baidu~\cite{LiuDBH15} & 1 & 1.3M & 99.13 & - \\
Center Face~\cite{wen2016} & 1 & 0.7M & 99.28 & 94.9 \\
Marginal loss~\cite{Deng_8014985} & 1 & 4M & 99.48 & 95.98 \\
Noisy Softmax~\cite{ChenDD17} & 1 & WebFace+ & 99.18 & 94.88 \\
CoCo loss~\cite{liu_2017_coco_v2} & 1 & 2M & \textbf{99.86} & - \\
\hline \hline
Range loss~\cite{Zhang_8237840} & 1 & 1.5M & 99.52 & 93.7 \\
Augmentation~\cite{2016_dowe} & 1 & WebFace & 98.06 & - \\
Center invariant loss~\cite{Wu_2017} & 1 & WebFace & 99.12 & 93.88 \\
Feature transfer~\cite{2018arXiv180309014Y} & 1 & 4.8M & 99.37 & - \\
\hline \hline
Softmax loss & 1 & WebFace & 97.88 & 93.1 \\
Softmax+Contrastive~\cite{Yi14} & 1 & WebFace & 98.78 & 93.5 \\
Triplet loss~\cite{Schroff15} & 1 & WebFace & 98.70 & 93.4 \\
L-Softmax loss~\cite{Liu2016} & 1 & WebFace & 99.10 & 94.0 \\
Softmax+Center loss~\cite{wen2016} & 1 & WebFace & 99.05 & 94.4 \\
SphereFace (A-Softmax)~\cite{liu2017sphereface} & 1 & WebFace & 99.42 & 95.0 \\
CosFace (LMCL)~\cite{2018Wang} & 1 & WebFace & 99.33 & 96.1 \\
\hline \hline
LMLE~\cite{huang2016lmle} & 1 & WebFace & 99.51 & 95.8 \\
CLMLE & 1 & WebFace & \textbf{99.62} & \textbf{96.5} \\
\hline
\end{tabular}
}
\label{tb2}
\end{table}

\noindent
{\bf Experiments on LFW and YTF.} Table~\ref{tb2} lists the face verification accuracy on the two datasets. Existing state-of-the-art face verification systems (in the first cell) either use large training data or model ensemble. For example, both the top performing CoCo loss~\cite{liu_2017_coco_v2} on LFW (99.86\%) and VGG Face~\cite{Parkhi15} on YTF (97.3\%) use more than 2M training images (the training set is regarded as \textit{small} only if it contains no more than 0.5M images). While the proposed method only uses the small, publicly available training data (CASIA-WebFace~\cite{YiLLL14a} with 0.45M images) and a single model. Our CLMLE achieves the best performance in this setting - 99.62\% on LFW and 96.5\% on YTF. By taking into account the class imbalance during training, our single CLMLE model even outperforms or performs closely to the FaceNet~\cite{Schroff15} which uses around 200M outside training data, and DeepID2+~\cite{SunWT15} which ensembles 25 models.

For a fair comparison with recent loss functions, we use the same WebFace training data and the same 64-layer CNN architecture. The compared loss functions are trained with their default hyper-parameters. One can observe that our CLMLE considerably outperforms the Softmax variants (including L-Softmax~\cite{Liu2016}, A-Softmax~\cite{liu2017sphereface}, and LMCL~\cite{2018Wang}) and metric learning methods (including contrastive~\cite{Yi14}, triplet~\cite{Schroff15} and center loss~\cite{wen2016}). They all suffer from the unimodal assumption of class distribution which is not suitable in class-imbalanced scenarios.

When compared with the recent imbalanced learning methods trained on WebFace or larger data (second cell in Table~\ref{tb2}), CLMLE is shown to achieve consistent gains. The methods of minority class augmentation in image space~\cite{2016_dowe} and feature space~\cite{2018arXiv180309014Y} obtain sub-optimal results, while the range loss~\cite{Zhang_8237840} and center invariant loss~\cite{Wu_2017} are not as discriminative as our CLMLE. Our approach demonstrates superior feature discrimination by enforcing large margins between local clusters from imbalanced classes. Moreover, CLMLE improves over its previous version LMLE~\cite{huang2016lmle} by discriminating entire cluster distributions rather than individual quintuplets sampled from them. Our large margin feature learning method is particularly suitable for generalization test in the open-set face recognition problem, as verified by the above experiments.

\begin{table}[t]
\caption{Face recognition on MegaFace Challenge 1~\cite{kemelmacher2016megaface} under the protocols of small and large training set. ``Rank 1'' refers to rank-1 face identification accuracy (\%) with 1M distractors, and ``Veri.'' refers to face verification TAR (\%) under $10^{-6}$ FAR. The state-of-the-art loss functions in the second-to-last cell and imbalanced learning loss functions in the last cell use the small training data (CASIA-WebFace~\cite{YiLLL14a}: 0.45M) and the same 64-layer CNN model for fair comparison.}
\centering
\resizebox{0.9\linewidth}{!}{
\begin{tabular}{ c|c|c|c }
\hline
Method & Protocol & Rank 1 & Veri. \\
\hline \hline
Beijing FaceAll\_Norm\_1600 & Large & 64.80 & 67.11 \\
Google - FaceNet v8 & Large & 70.49 & 86.47 \\
NTechLAB - facenx\_large & Large & 73.30 & 85.08 \\
SIATMMLAB TencentVision & Large & 74.20 & 87.27 \\
DeepSense V2 & Large & 81.29 & \textbf{95.99} \\
YouTu Lab & Large & 83.29 & 91.34 \\
Vocord - deepVo V3 & Large & \textbf{91.76} & 94.96 \\
\hline \hline
SIAT\_MMLAB & Small & 65.23 & 76.72 \\
DeepSense - Small & Small & 70.98 & 82.85 \\
SphereFace - Small & Small & 75.76 & 90.04 \\
Beijing FaceAll V2 & Small & 76.66 & 77.60 \\
GRCCV & Small & 77.67 & 74.88 \\
FUDAN-CS\_SDS & Small & 77.98 & 79.19 \\
CoCo loss~\cite{liu_2017_coco_v2} & Small & 76.57 & - \\
\hline \hline
Softmax loss & Small & 54.85 & 65.92 \\
Softmax+Contrastive~\cite{Yi14} & Small & 65.21 & 78.86 \\
Triplet loss~\cite{Schroff15} & Small & 64.79 & 78.32 \\
L-Softmax loss~\cite{Liu2016} & Small & 67.12 & 80.42 \\
Softmax+Center loss~\cite{wen2016} & Small & 65.49 & 80.14 \\
SphereFace (A-Softmax)~\cite{liu2017sphereface} & Small & 72.72 & 85.56 \\
CosFace (LMCL)~\cite{2018Wang} & Small & 77.11 & 89.88 \\
\hline \hline
Range loss~\cite{Zhang_8237840} & Small & 72.94 & 83.62 \\
LMLE~\cite{huang2016lmle} & Small & 78.53 & 89.45 \\
CLMLE & Small & \textbf{79.68} & \textbf{91.85} \\
\hline
\end{tabular}
}
\label{tb3}
\end{table}

\noindent
{\bf Experiments on MegaFace Challenge 1.} The feature generalization can be further tested in the MegaFace Challenge 1~\cite{kemelmacher2016megaface} where the gallery set contains more than 1M distractors. For the face identification task which aims to match one probe image to the images with the same person in the gallery, the large MegaFace gallery set poses a big challenge and naturally causes data imbalance. For face verification, we should decide if an image pair contains the same person or not. There are 4 billion negative pairs generated between the probe and gallery sets, which are imbalanced with respect to the available positive pairs.

Table~\ref{tb3} first reports the face identification and verification results in terms of the simple rank-1 identification accuracy and verification True Accept Rate (TAR) at fixed False Accept Rate (FAR). Our training set CASIA-WebFace~\cite{YiLLL14a} has 0.45M images, thus we follow the small training set protocol. Under this protocol, our CLMLE performs best for both the identification and verification tasks. For some models trained under the large training set protocol,~\eg,~Google-FaceNet v8 with 500M images and NTechLAB-facenx\_large with 18M images, they are also beaten by our CLMLE with small training data. In comparison to the recent models trained with the same small data and network architecture (in second-to-last cell), our CLMLE shows better results and superior generalization ability again. We further implemented the range loss~\cite{Zhang_8237840} (a reportedly competitive imbalanced learning method) and LMLE~\cite{huang2016lmle} in the same settings. For the range loss, we formed a weighted combination of it and softmax loss, with weights suggested by the authors. It is evident from the table that CLMLE outperforms in the imbalance handling ability.

\begin{figure}[t]
\begin{center}
\includegraphics[width=1.0\linewidth]{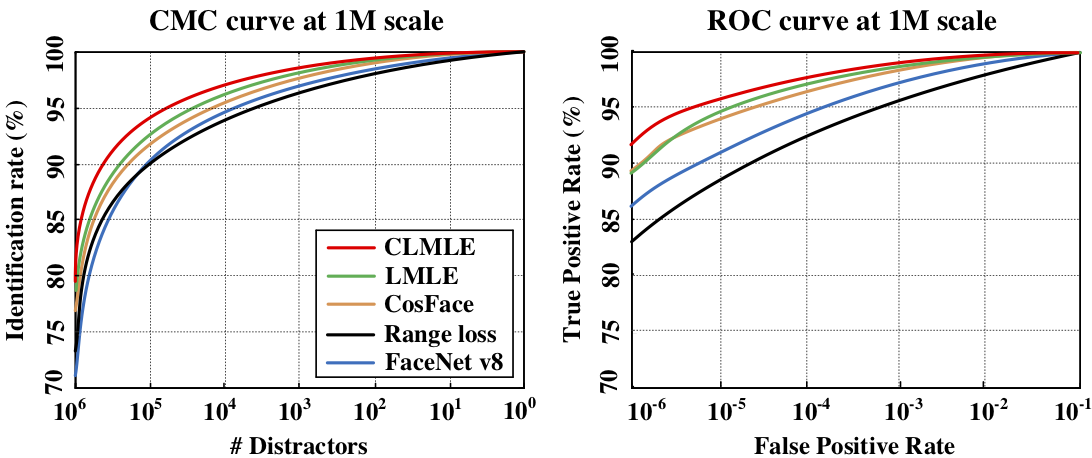}
\end{center}
\vspace{-1.5em}
\caption{CMC and ROC curves of top performing methods with 1M distractors on MegaFace Challenge 1~\cite{kemelmacher2016megaface}. Note the FaceNet v8 follows the large training set protocol, while other methods follow the small one.}
\vspace{-1em}
\label{fig7}
\end{figure}

\begin{figure}[t]
\begin{center}
\includegraphics[width=1.0\linewidth]{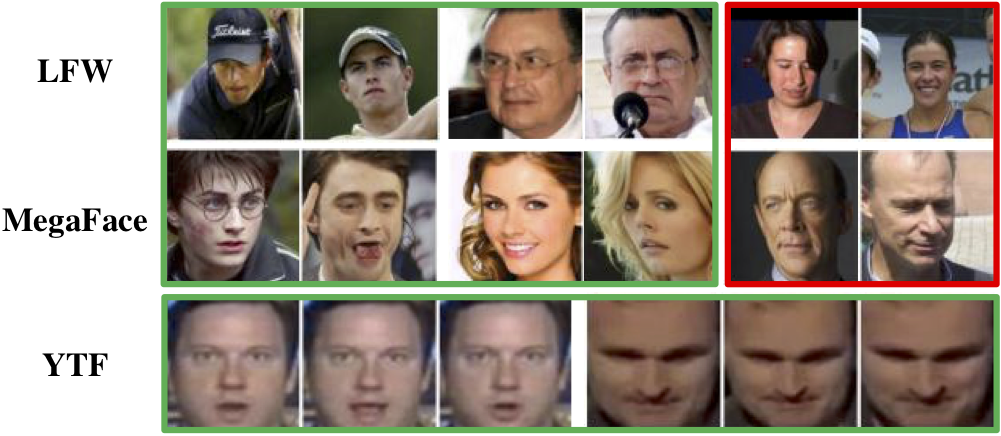}
\end{center}
\vspace{-1.5em}
\caption{Challenging pairs (green: positive pair; red: negative pair) that are correctly recognized by our method.}
\vspace{-0.8em}
\label{fig8}
\end{figure}

\begin{table}[!t]
\caption{Face recognition on MegaFace Challenge 2~\cite{Nech_2017_CVPR} under the large training set protocol. ``Rank 1'' refers to rank-1 face identification accuracy (\%) with 1M distractors, and ``Veri.'' refers to face verification TAR (\%) under $10^{-6}$ FAR. The SphereFace~\cite{liu2017sphereface}, CosFace~\cite{2018Wang} and imbalanced learning methods in the last cell use the same 64-layer CNN model for fair comparison.}
\centering
\resizebox{0.9\linewidth}{!}{
\begin{tabular}{ c|c|c|c }
\hline
Method & Protocol & Rank 1 & Veri. \\
\hline \hline
3DiVi & Large & 57.04 & 66.45 \\
NEC & Large & 62.12 & 66.84 \\
GRCCV & Large & \textbf{75.77} & 74.84 \\
SphereFace (A-Softmax)~\cite{liu2017sphereface} & Large & 71.17 & 84.22 \\
CosFace (LMCL)~\cite{2018Wang} & Large & 74.11 & \textbf{86.77} \\
\hline \hline
Range loss~\cite{Zhang_8237840} & Large & 69.54 & 82.67 \\
LMLE~\cite{huang2016lmle} & Large & 74.76 & 87.78 \\
CLMLE & Large & \textbf{76.26} & \textbf{89.41} \\
\hline
\end{tabular}
}
\label{tb4}
\end{table}

Fig.~\ref{fig7} shows the Cumulative Match Characteristics (CMC) curves (for face identification) as well as the Receiver Operating Characteristic (ROC) curves (for verification) that can take into account any imbalance in testing pairs. Note the MegaFace gallery set contains different scales of distractors, from 10 to 1M with increasing difficulty. The figure shows the curves at 1M scale for the top performing methods under small and large protocols. We can see that our CLMLE method achieves the new state-of-the-art performance (see Fig.~\ref{fig8} for some visual results).

\noindent
{\bf Experiments on MegaFace Challenge 2.} For MegaFace Challenge 2~\cite{Nech_2017_CVPR}, all the algorithms use the same training data provided by MegaFace and follow the large training set protocol. The training set includes 4.7M images from 672k identities. The gallery set contains 1M images that are different from those in Challenge 1. Table~\ref{tb4} illustrates our CLMLE attains the top performance for both face identification (76.26\% rank-1 accuracy) and face verification (89.41\% TAR under $10^{-6}$ FAR). This validates the importance of tackling imbalance for feature representation learning, and our CLMLE outperforms its previous version LMLE and the range loss~\cite{Zhang_8237840} for imbalanced learning. Moreover, our large margin nature is shown to significantly improve the open-set recognition performance. By contrast, the range loss as a regularization method over softmax, is more suitable to closed-set classification problems.

\subsection{Face Attribute Prediction}

Table~\ref{tb5} compares our CLMLE method for face attribute prediction with the state-of-the-art methods of Triplet-kNN~\cite{Schroff15}, PANDA~\cite{Ning14} and ANet~\cite{liu15} which are trained on the same data and tuned to their best performance. The attributes and their mean per-class accuracies are given in the ascending order of class imbalance level ($=|$positive class rate-50$|$\%). This is to highlight the impact of class imbalance on performance. Note the CelebA dataset~\cite{liu15} also poses challenges to joint attribute prediction, where the variant imbalance levels of different attributes need to be handled simultaneously.

\begin{table*}[t]
\caption{Mean per-class accuracy (\%) and class imbalance level ($=|$positive class rate-50$|$\%) of each of the 40 binary attributes on CelebA dataset~\cite{liu15}. Attributes are sorted by the imbalance level in an ascending order. The second and third cells list state-of-the-art methods and imbalanced learning methods, respectively. Note the results of balanced accuracy are different from the classification accuracy reported by ANet~\cite{liu15}. Also, MOON (-D) and AFFACT (-D) use networks of VGG-16 and ResNet-50, much larger than ours as described in~\cite{Yi14}. The suffix 'D' denotes domain adaptation to the balanced attribute distribution, the underlying distribution suggested by our balanced evaluation metric.
}
\centering
\resizebox{0.95\textwidth}{!}{
\begin{tabular}{c|c|c|c|c|c|c|c|c|c|c|c|c|c|c|c|c|c|c|c|c|c}
\hline
 & \rotatebox[origin=lB]{90}{Attractive}
 & \rotatebox[origin=lB]{90}{Mouth Open}
 & \rotatebox[origin=lB]{90}{Smiling}
 & \rotatebox[origin=lB]{90}{Wear Lipstick}
 & \rotatebox[origin=lB]{90}{High Cheekbones}
 & \rotatebox[origin=lB]{90}{Male}
 & \rotatebox[origin=lB]{90}{Heavy Makeup}
 & \rotatebox[origin=lB]{90}{Wavy Hair}
 & \rotatebox[origin=lB]{90}{Oval Face}
 & \rotatebox[origin=lB]{90}{Pointy Nose}
 & \rotatebox[origin=lB]{90}{Arched Eyebrows}
 & \rotatebox[origin=lB]{90}{Black Hair}
 & \rotatebox[origin=lB]{90}{Big Lips}
 & \rotatebox[origin=lB]{90}{Big Nose}
 & \rotatebox[origin=lB]{90}{Young}
 & \rotatebox[origin=lB]{90}{Straight Hair}
 & \rotatebox[origin=lB]{90}{Brown Hair}
 & \rotatebox[origin=lB]{90}{Bags Under Eyes}
 & \rotatebox[origin=lB]{90}{Wear Earrings}
 & \rotatebox[origin=lB]{90}{No Beard}
 & \rotatebox[origin=lB]{90}{Bangs} \\
\hline \hline
Imbalance level  & 1 & 2 & 2 & 3 & 5 & 8 & 11 & 18 & 22 & 22 & 23 & 26 & 26 & 27 & 28 & 29 & 30 & 30 & 31 & 33 & 35 \\ \hline
Triplet-kNN~\cite{Schroff15} & 83 & 92 & 92 & 91 & 86 & 91 & 88 & 77 & 61 & 61 & 73 & 82 & 55 & 68 & 75 & 63 & 76 & 63 & 69 & 82 & 81 \\
 PANDA~\cite{Ning14} & 85 & 93 & 98 & 97 & 89 & 99 & 95 & 78 & 66 & 67& 77 & 84 & 56 & 72 & 78 & 66 & 85 & 67 & 77 & 87 & 92 \\
 ANet~\cite{liu15} &87 & 96 & 97 & 95 & 89 & 99 & 96 & 81 & 67  &69 &76 &90 &57&  78  &84 &69 &83 &70&  83  &93 &90 \\ \hline
Down-sampling~\cite{Drummond03} &78 & 87 & 90 & 91 & 80 & 90 & 89 & 70 & 58  &63 &70 &80 &61&  76  &80 &61 &76 &71&  70  &88 &88 \\
MOON~\cite{Rudd2016} & 82 & 94 & 93 & 94 & 87 & 98 & 91 & 79 & 65 & 66 & 78 & 86 & 59 & 72 & 79 & 68 & 81 & 70 & 80 & 88 & 92 \\
Over-sampling~\cite{Drummond03} &77 & 89 & 90 & 92 & 84 & 95 & 87 & 70 & 63  &67 &79 &84 &61&  73  &75 &66 &82 &73&  76  &88 &90 \\ 
Cost-sensitive~\cite{He09} &78 & 89 & 90 & 91 & 85 & 93 & 89 & 75 & 64  &65 &78 &85 &61&  74  &75 &67 &84 &74&  76  &88 &90 \\
AFFACT~\cite{guenther2017affact} & 83 & 94 & 93 & 94 & 88 & 98 & 91 & 84 & 66 & 66 & 80 & 87 & 62 & 76 & 82 & 75 & 83 & 76 & 86 & 94 & 92 \\
CRL-I~\cite{Dong_2017_ICCV} &83 & 95 & 93 & 94 & 89 & 96 & 84 & 79 & 66  &73 &80 &90 &68&  80  &84 &73 &86 &80&  83  &94 &95 \\
MOON-D~\cite{Rudd2016} & 82 & 94 & 93 & 94 & 87 & 98 & 91 & 81 & 69 & 71 & 82 & 88 & 66 & 77 & 84 & 75 & 85 & 81 & 86 & 94 & 94 \\
AFFACT-D~\cite{guenther2017affact} & 83 & 94 & 93 & 94 & 88 & 98 & 92 & 86 & 72 & 72 & 84 & 89 & 69 & 79 & 86 & 82 & 86 & 83 & 89 & 95 & 95 \\\hline
LMLE~\cite{huang2016lmle} & 88&  96& 99& 99& 92& 99& 98& 83& 68  &72&  79& 92& 60& 80& 87& 73& 87& 73& 83& 96& 98  \\
CLMLE & 90&97&99&98&94&99&98&87&72&78&86&95&66&85&90&80&89&82&86&98&99 \\
\hline \hline
 & \rotatebox[origin=lB]{90}{Blond Hair}
 & \rotatebox[origin=lB]{90}{Bushy Eyebrows}
 & \rotatebox[origin=lB]{90}{Wear Necklace}
 & \rotatebox[origin=lB]{90}{Narrow Eyes}
 & \rotatebox[origin=lB]{90}{5 o'clock Shadow}
 & \rotatebox[origin=lB]{90}{Receding Hairline}
 & \rotatebox[origin=lB]{90}{Wear Necktie}
 & \rotatebox[origin=lB]{90}{Eyeglasses}
 & \rotatebox[origin=lB]{90}{Rosy Cheeks}
 & \rotatebox[origin=lB]{90}{Goatee}
 & \rotatebox[origin=lB]{90}{Chubby}
 & \rotatebox[origin=lB]{90}{Sideburns}
 & \rotatebox[origin=lB]{90}{Blurry}
 & \rotatebox[origin=lB]{90}{Wear Hat}
 & \rotatebox[origin=lB]{90}{Double Chin}
 & \rotatebox[origin=lB]{90}{Pale Skin}
 & \rotatebox[origin=lB]{90}{Gray Hair}
 & \rotatebox[origin=lB]{90}{Mustache}
 & \rotatebox[origin=lB]{90}{Bald}
 &
 & \rotatebox[origin=lB]{90}{\textbf{Average}} \\
\hline \hline
Imbalance level & 35 & 36 & 38 & 38 & 39 & 42 & 43 & 44 & 44 & 44 & 44 & 44 & 45 & 45 & 45 & 46 & 46 & 46 & 48 &    &  \\ \hline
Triplet-kNN~\cite{Schroff15} & 81&  68& 50& 47& 66& 60& 73& 82& 64& 73& 64& 71& 43& 84& 60& 63& 72& 57& 75 &    & 71.55 \\
 PANDA~\cite{Ning14} & 91&  74& 51& 51& 76& 67& 85& 88& 68& 84& 65& 81& 50& 90& 64& 69& 79& 63& 74  &    & 76.95 \\
 ANet~\cite{liu15} & 90&  82& 59& 57& 81& 70& 79& 95& 76& 86& 70& 79& 56& 90& 68& 77& 85& 61& 73&    & 79.58 \\ \hline
Down-sampling~\cite{Drummond03} & 85&  75& 66& 61& 82& 79& 80& 85& 82& 85& 78& 80& 68& 90& 80& 78& 88& 60&  79&  & 77.45 \\
MOON~\cite{Rudd2016} & 91 & 79 & 57 & 58 & 78 & 69 & 82 & 97 & 70 & 84 & 66 & 82 & 71 & 91 & 66 & 69 & 81 & 68 & 83 & & 78.59 \\
Over-sampling~\cite{Drummond03} & 90&  80& 71& 65& 85& 82& 79& 91& 90& 89& 83& 90& 76& 89& 84& 82& 90& 90&  92&  & 81.48 \\
Cost-sensitive~\cite{He09}      & 89&  79& 71& 65& 84& 81& 82& 91& 92& 86& 82& 90& 76& 90& 84& 80& 90& 88&  93&  & 81.60 \\
AFFACT~\cite{guenther2017affact} & 91 & 80 & 70 & 64 & 86 & 76 & 90 & 99 & 79 & 90 & 76 & 91 & 74 & 95 & 74 & 75 & 87 & 74 & 89 & & 82.69 \\ 
CRL-I~\cite{Dong_2017_ICCV}     & 95&  84& 74& 72& 90& 87& 88& 96& 88& 96& 87& 92& 85& 98& 89& 92& 95& 94&  97&  & 86.60 \\
MOON-D~\cite{Rudd2016} & 94 & 86 & 76 & 75 & 90 & 88 & 93 & 99 & 89 & 94 & 87 & 95 & 86 & 96 & 88 & 91 & 94 & 93 & 96 & & 87.02 \\
AFFACT-D~\cite{guenther2017affact} & 94 & 87 & 81 & 79 & 92 & 88 & 94 & 99 & 91 & 96 & 89 & 95 & 91 & 98 & 90 & 92 & 95 & 95 & 97 & & \textbf{88.84} \\\hline
LMLE~\cite{huang2016lmle} & 99& 82& 59& 59& 82& 76& 90& 98& 78& 95& 79& 88& 59& 99& 74& 80& 91& 73& 90  &    & 83.83 \\
CLMLE & 99&88&69&71&91&82&96&99&86&98&85&94&72&99&87&94&96&82&95 & & \textbf{88.78} \\ \hline
\end{tabular}
}
\label{tb5}
\vspace{-1em}
\end{table*}

\begin{figure}[t]
\begin{center}
\includegraphics[width=1.0\linewidth]{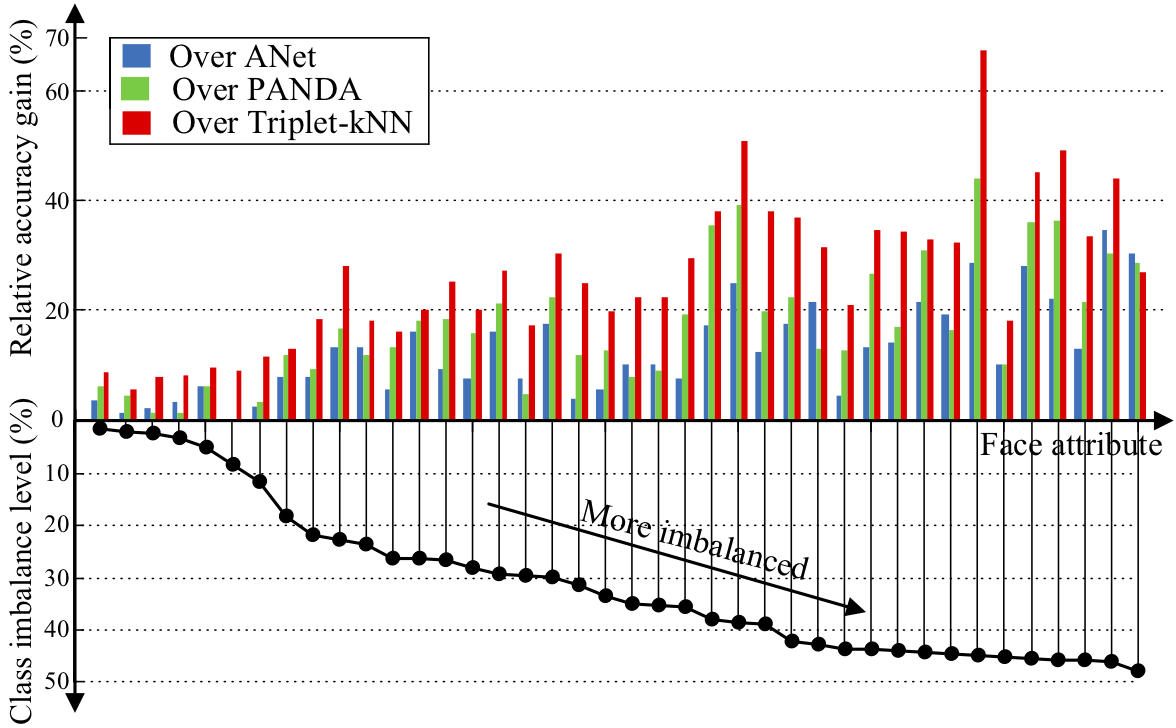}
\end{center}
\vspace{-1.5em}
\caption{Relative gains over the state-of-the-art methods without imbalance handling mechanism across the 40 imbalanced attributes on CelebA~\cite{liu15}.}
\label{fig9}
\end{figure}

It is shown that CLMLE outperforms these methods across all attributes, with an average gap of about 9\% over ANet. Considering most face attributes exhibit high class imbalance with an average positive class rate of only 23\%, such improvements are nontrivial and prove the efficacy of our learned features on imbalanced data. Although PANDA and ANet are capable of learning robust feature representations by ensemble and multi-task learning respectively, they ignore the imbalance issue and thus struggle for highly imbalanced attributes, \eg,~``Bald''. By contrast, our method performs well on such attributes. Our advantage over these non-imbalanced learning methods is more evident when observing the relative accuracy gains for different attributes. Fig.~\ref{fig9} shows the gains tend to be larger for those attributes with a higher class imbalance level.

The bottom cells of Table~\ref{tb5} illustrate the results of imbalanced learning methods. We can see that our CLMLE outperforms traditional re-sampling and cost-sensitive learning techniques, as well as the previous LMLE method~\cite{huang2016lmle} and recent CRL-I~\cite{Dong_2017_ICCV}, all of which use the same network architecture~\cite{Yi14} for fair comparison. CRL-I denotes regularizing the minority class with Instance level hard mining. However it still cannot guarantee equal learning for all classes and has inferior results. Thanks to the authors of MOON~\cite{Rudd2016} (with adaptive re-weighting) and AFFACT~\cite{guenther2017affact}, we further compare against their provided results of balanced accuracy from using VGG-16 and ResNet-50 networks respectively. Our CLMLE still outperforms the two methods by large margin (about 10\% and 6\% on average), even using a much smaller network~\cite{Yi14}. On top of larger networks, the MOON-D and AFFACT-D variants are further fine-tuned to completely balanced attribute distribution, the same as the underlying distribution suggested by our balanced evaluation metric. While CLMLE, without assuming the target attribute distribution or using large networks, is able to achieve comparable or even better performance.

\begin{figure}[t]
\begin{center}
\includegraphics[width=1.0\linewidth]{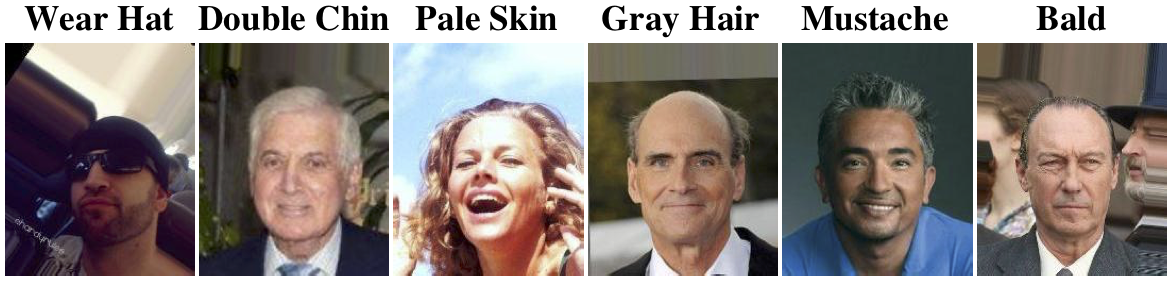}
\end{center}
\vspace{-1.5em}
\caption{Most imbalanced face attributes (from Table~\ref{tb5}) that are correctly predicted by our method.}
\label{fig10}
\end{figure}

Table~\ref{tb6} compares CLMLE with state-of-the-art AttCNN~\cite{AAAI1816962} and SSP~\cite{Kalayeh17} in terms of two evaluation metrics. AttCNN addresses the multi-label imbalance problem by selective data learning. However, only average classification accuracy is reported, and our superiority still holds for this metric. The Semantic Segmentation-based Pooling (SSP) method does not handle imbalance, but adds a segmentation network to improve the attribute network. Our CLMLE is quite comparable in both metrics without increasing network capacity. Fig.~\ref{fig10} shows some challenging images with highly imbalanced attributes that are correctly predicted by CLMLE but not by SSP.

\begin{table}[t]
\caption{Average balanced accuracy and classification accuracy (\%) for the 40 binary attributes on CelebA dataset~\cite{liu15}.}
\centering
\resizebox{\linewidth}{!}{
\begin{tabular}{ c|c|c|c }
\hline
Method & AttCNN~\cite{AAAI1816962} & SSP~\cite{Kalayeh17} & CLMLE \\
\hline
\hline
Average balanced acc. & - & 88.24 & \textbf{88.78} \\
Average classification acc. & 90.97 & \textbf{91.67} & 91.13 \\
\hline
\end{tabular}
}
\label{tb6}
\end{table}

\begin{table}[t]
\caption{Ablation study in terms of the face verification accuracy (\%) on LFW and average balanced accuracy (\%) for attribute prediction on CelebA.}
\centering
\resizebox{0.7\linewidth}{!}{
\begin{tabular}{ c|c|c }
\hline
Method & LFW & CelebA \\
\hline
\hline
CLMLE & \textbf{99.62} & \textbf{88.78} \\ \hline
uniform cluster sampling & 99.52 & 88.23 \\
no cost-sensitive learning & 99.57 & 87.46 \\
\hline
\end{tabular}
}
\label{tb7}
\end{table}

\begin{table}[!t]
\caption{Mean class-balanced accuracy (\%) on CIFAR-100~\cite{Krizhevsky09} and MNIST-rot-back-image~\cite{Larochelle07} datasets with simulated class imbalance (with parameter $\gamma$ - the smaller, the more imbalanced). Triplet+ denotes triplet loss~\cite{Schroff15} with over-sampling and cost sensitivity. Balanced baseline denotes the CE+CRL results under class-balanced setting.}
\centering
\resizebox{1.0\linewidth}{!}{
\begin{tabular}{ c|c|c|c|c }
\hline
Method & CIFAR-100 & CIFAR-100 & MNIST & MNIST \\
       & $\gamma=1$ & $\gamma=0.5$ & $\gamma=1$ & $\gamma=0.5$ \\
\hline
\hline
Triplet+ & 42.2 & 32.2 & 65.2 & 59.8 \\
CE+CRL~\cite{Dong_2017_ICCV} & 45.2 & 37.4 & 71.2 & 64.5 \\ \hline
LMLE~\cite{huang2016lmle} & 44.3 & 38.1 & 71.9 & 65.7 \\
CLMLE & \textbf{47.4} & \textbf{39.7} & \textbf{73.5} & \textbf{68.8} \\ \hline
Balanced baseline & \multicolumn{2}{c|}{48.2} & \multicolumn{2}{c}{78.3} \\
\hline
\end{tabular}
}
\label{tb8}
\end{table}

\subsection{Ablation Study}
Table~\ref{tb7} quantifies the benefits of the re-sampling and cost-sensitive components adopted in our CLMLE method. Both the classic schemes prove themselves as useful for our imbalanced tasks. The sampling of query cluster $\mathcal{I}_1$ is designed to be non-uniform in class - we pick the one with the highest observed loss. Results show that it hurts performance if we choose uniform cluster sampling instead. It is more so for face recognition, because compared to the binary attribute prediction problem, face recognition has far more classes whose discrimination is more sensitive to the hard modes (clusters). The cost-sensitive scheme is used by us to re-balance the classes in batch at score level. Note when this scheme is applied to perfectly class-balanced batches, it would have no effects. However in the case of multi-label attributes, class-balanced batch samples for one attribute will almost certainly be imbalanced for the other attributes. This is when the class costs can help. Our ablation study confirms that the cost-sensitivity has a larger impact on CelebA attributes. To highlight the effects of our \textit{k}-nearest cluster classifier in Eq.~(\ref{eq10}), we replace it with the regular instance-wise kNN classifier. As expected, we observe much lower speed and a bit worse results (88.59\%) for attributes.

\subsection{Generic Imbalanced Image Classification}

In addition to the face recognition and attribute prediction tasks with real-world imbalanced data, we have also tested CLMLE on generic image classification problems (single-label \& multi-class). We experiment on standard benchmarks CIFAR-100~\cite{Krizhevsky09} and MNIST-rot-back-image~\cite{Larochelle07} but with simulated class imbalance. As in~\cite{Dong_2017_ICCV}, the simulated class-imbalanced data follow a power-law distribution $f(c)=L_c^{\max}/(c^{\gamma}+L_c^{\min})$ where $L_c^{\max}$ and $L_c^{\min}$ are the largest and smallest sizes of classes $c\in[1,C]$. The parameter $\gamma$ controls the degree of class imbalance --- the smaller $\gamma$ is, the more imbalanced the distribution is. We follow the same experimental settings of~\cite{Dong_2017_ICCV} and~\cite{huang2016lmle} for CIFAR-100 and MNIST respectively, including data usage, network architecture and hyper-parameters. Table~\ref{tb8} reports classification results under two imbalnaced ratios $\gamma=\{1,0.5\}$, and compares CLMLE with some imbalanced learning baselines Triplet+, CE (Cross-Entropy)+CRL~\cite{Dong_2017_ICCV} and LMLE~\cite{huang2016lmle}. Classification results under the class-balanced scenario are reported using CE+CRL as a balanced baseline. We can see that CLMLE's superiority still holds for imbalanced image classification.

\section{Conclusion}
Class imbalance is common in many vision tasks, including face recognition and attribute prediction. Contemporary deep representation learning methods typically adopt class re-sampling or cost-sensitive learning schemes. Through extensive experiments, we have validated their usefulness and further demonstrated that the proposed Cluster-based Large Margin Local Embedding (CLMLE) works remarkably well when just combined with a simple \textit{k}-nearest cluster classifier. CLMLE maintains inter-cluster angular margins both within and between classes, thus carving much more balanced class boundaries locally. Our feature learning converges fast and achieves the new state-of-the-art performance on existing face recognition and attribute benchmarks.

\ifCLASSOPTIONcompsoc
  \section*{Acknowledgments}
\else
  \section*{Acknowledgment}
\fi

This work is supported by SenseTime Group Limited and the General Research Fund sponsored by the Research Grants Council of the Hong Kong SAR (CUHK 14241716, 14224316, 14209217).

\ifCLASSOPTIONcaptionsoff
  \newpage
\fi


\bibliographystyle{IEEEtran}
\bibliography{myref}

\begin{thebibliography}{10}
\bibitem{AAAI1816962}
E.~Hand, C.~Castillo, and R.~Chellappa, ``Doing the best we can with what we
  have: Multi-label balancing with selective learning for attribute
  prediction,'' in \emph{AAAI}, 2018.
\end{thebibliography}

%

\begin{IEEEbiography}[{\includegraphics[width=1in,height=1.25in,clip,keepaspectratio]{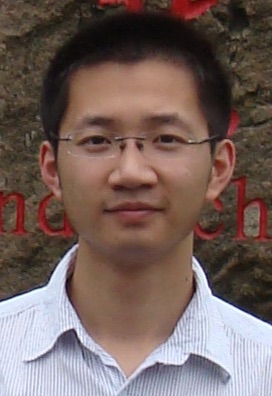}}]{Chen Huang}
received the Ph.D. degree in Electronic Engineering from Tsinghua University, Beijing, China, in 2014. He was a postdoctoral fellow in the Robotics Institute of Carnegie Mellon University, and also in the Department of Information Engineering, Chinese University of Hong Kong. His research interests include machine learning and computer vision, with focus on deep learning and face analysis.
\end{IEEEbiography}


\begin{IEEEbiography}[{\includegraphics[width=1in,height=1.25in,clip,keepaspectratio]{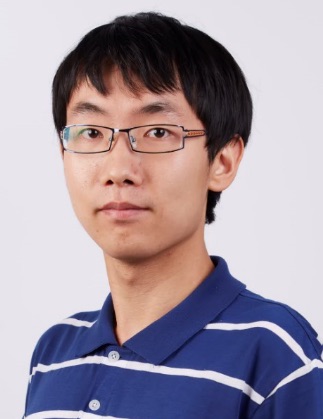}}]{Yining Li}
received the B.E degree in Automation from Tsinghua University in 2014. He is currently working toward the Ph.D. degree in the Department of Information Engeneering, Chinese University of Hong Kong. His research interests include computer vision and machine learning, with focus on semantic attribute recognition and analysis.
\end{IEEEbiography}

\begin{IEEEbiography}[{\includegraphics[width=1in,height=1.25in,clip,keepaspectratio]{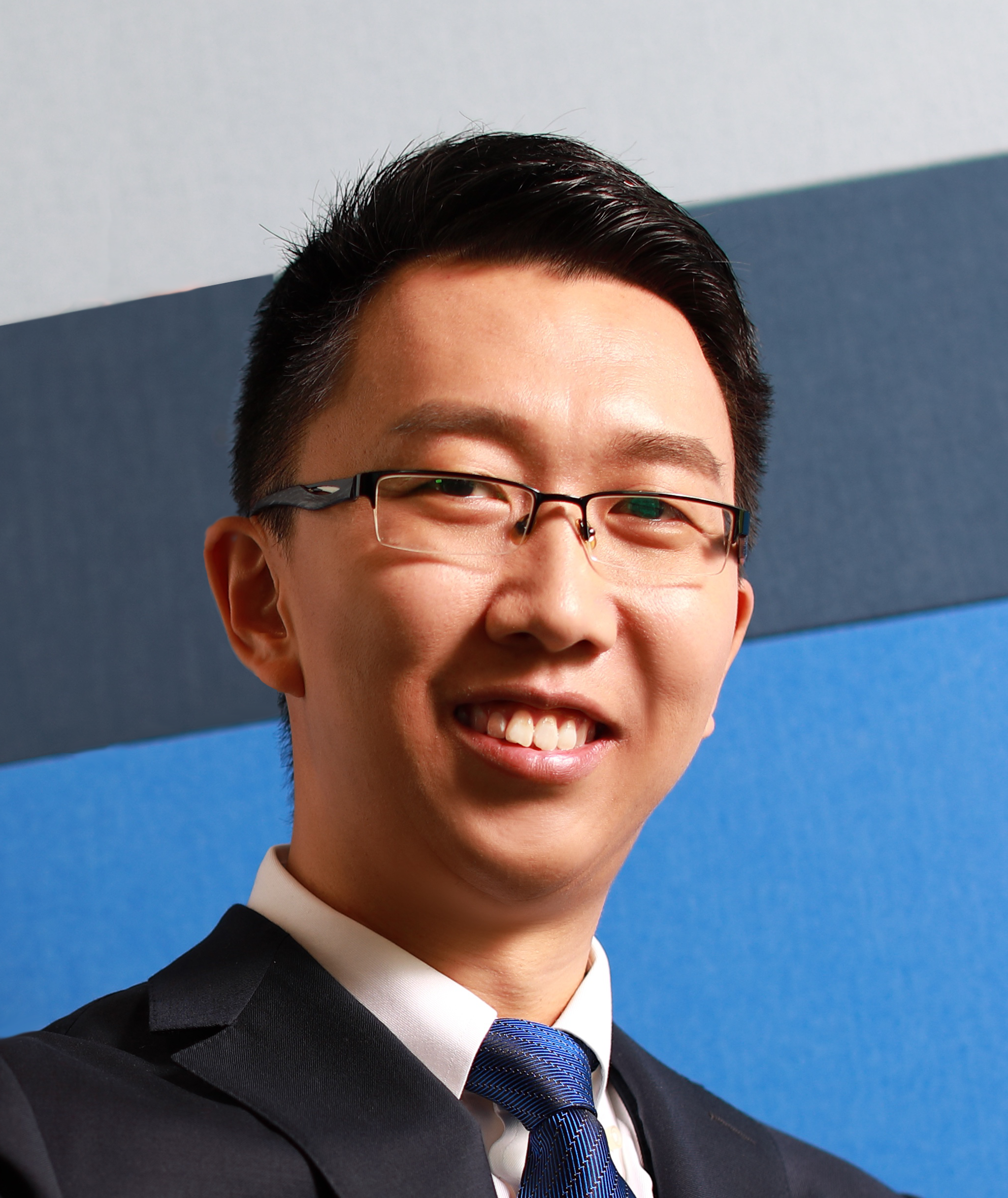}}]{Chen Change Loy}
(S'06-M'10-SM'17) received the Ph.D. degree in computer science from the Queen Mary University of London in 2010. He is an Associate Professor with the School of Computer Science and Engineering, Nanyang Technological University. Prior to joining NTU, he served as a Research Assistant Professor with the Department of Information Engineering, The Chinese University of Hong Kong, from 2013 to 2018. His research interests include computer vision and deep learning, with focus on face analysis, image processing, and visual surveillance. He serves as an Associate Editor of the International Journal of Computer Vision. He also serves/served as an Area Chair of CVPR 2019, BMVC 2019, ECCV 2018 and BMVC 2018. He is a senior member of IEEE.
\end{IEEEbiography}

\begin{IEEEbiography}[{\includegraphics[width=1in,height=1.25in,clip,keepaspectratio]{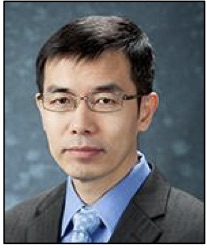}}]{Xiaoou Tang}
(S'93-M'96-SM'02-F'09) received
the BS degree from the University of Science
and Technology of China, Hefei, in 1990, the MS
degree from the University of Rochester, New
York, in 1991, and the PhD degree from the Massachusetts
Institute of Technology, Cambridge,
in 1996. He is a Professor of the Department of Information Engineering, The Chinese University of Hong Kong. He worked
as the group manager of the Visual Computing
Group at the Microsoft Research Asia, from 2005 to 2008. His research
interests include computer vision, pattern recognition, and video processing.
He received the Best Paper Award at the IEEE Conference
on Computer Vision and Pattern Recognition (CVPR) 2009 and Outstanding Student Paper Award at the AAAI 2015. He was a
program chair of the IEEE International Conference on Computer Vision
(ICCV) 2009 and served as an associate editor of the IEEE Transactions on
Pattern Analysis and Machine Intelligence. He is an Editor-in-Chief of the International Journal of Computer Vision. He is a fellow of the IEEE.
\end{IEEEbiography}


\vfill


\end{document}